\documentclass[10pt,twocolumn,letterpaper]{article}

\usepackage{iccv}              
\usepackage{xcolor}
\usepackage{booktabs}
\usepackage{multirow}
\usepackage{colortbl}
\usepackage{makecell}
\usepackage{graphicx}
\usepackage{subcaption}
\newcommand{\modelname}{ProSAM}
\newcommand{\best}[1]{\textcolor[HTML]{23bb17}{\textbf{#1}}}
\newcommand{\sbest}[1]{\textcolor[HTML]{4e4fd1}{\textbf{#1}}}

\newcommand{\cc}[1]{\cellcolor[HTML]{ededed}{#1}}

\newcommand{\ourrow}{\rowcolor[HTML]{ededed}}

\usepackage[pagebackref,breaklinks,colorlinks,allcolors=iccvblue]{hyperref}
\usepackage{bm}
\usepackage{arydshln}
\usepackage{colortbl}
\usepackage{amsmath, amsthm, amssymb}
\newtheorem{proposition}{Proposition}

\newtheorem{corollary}{Corollary}

\usepackage{minitoc}
\usepackage[toc,page,header]{appendix}

\definecolor{iccvblue}{rgb}{0.21,0.49,0.74}
\usepackage[pagebackref,breaklinks,colorlinks,allcolors=iccvblue]{hyperref}

\def\paperID{12134} 
\def\confName{ICCV}
\def\confYear{2025}

\title{ProSAM: Enhancing the Robustness of SAM-based Visual Reference Segmentation with Probabilistic Prompts}

\author{{
Xiaoqi Wang$^{1,2}$, Clint Sebastian$^2$, 
Wenbin He$^{1,2}$, Liu Ren$^{1,2}$
}\\
{
$^1$Bosch Research North America, $^2$Bosch Center for Artificial Intelligence (BCAI)
}\\
{\tt\small xiaoqi.wang@us.bosch.com clint.sebastian@de.bosch.com wenbin.he2@us.bosch.com liu.ren@us.bosch.com}
}

\begin{document}
\doparttoc 
\faketableofcontents 
\maketitle
\begin{abstract}
 The recent advancements in large foundation models have driven the success of open-set image segmentation, a task focused on segmenting objects beyond predefined categories. Among various prompt types (such as points, boxes, texts, and visual references), visual reference segmentation stands out for its unique flexibility and strong zero-shot capabilities. Recently, several SAM-based methods have made notable progress in this task by automatically generating prompts to guide SAM. However, these methods often generate prompts at boundaries of target regions due to suboptimal prompt encoder, which results in instability and reduced robustness. In this work, we introduce \modelname, a simple but effective method to address the stability challenges we identified in existing SAM-based visual reference segmentation approaches. By learning a variational prompt encoder to predict multivariate prompt distributions, \modelname\ avoids generating prompts that lie in unstable regions, overcoming the instability caused by less robust prompts. Our approach consistently surpasses state-of-the-art methods on the Pascal-5$^i$ and COCO-20$^i$ datasets, providing a more robust solution for visual reference segmentation.
\end{abstract}

\section{Introduction}
\label{sec:intro}

Open-set image segmentation methods have gained considerable attention for their ability to segment objects beyond a fixed set of categories. These methods incorporate diverse prompts, including points, boxes, texts, and visual references, effectively addressing the limitations of closed-set approaches. The introduction of the Segment Anything Model (SAM) series~\cite{sam,sam2} has notably advanced open-set segmentation performance using point and box prompts. However, when segmenting the same type of object across multiple images, using SAM can be tedious and time-consuming because it requires custom prompts for each image individually. Image segmentation methods using text prompts~\cite{san,use,clips4} offer a solution to this limitation but encounter two main challenges~\cite{trex2}: (i) aligning vision and language representations for rare or long-tailed objects is challenging due to their scarcity, leading to compromised segmentation performance for these objects; and (ii) certain objects are difficult to describe accurately in natural language without specialized knowledge. For example, someone without a background in chemistry may struggle to accurately describe "molecular orbitals".

\begin{figure}[tb]
  \centering
  \includegraphics[width=0.76\linewidth]{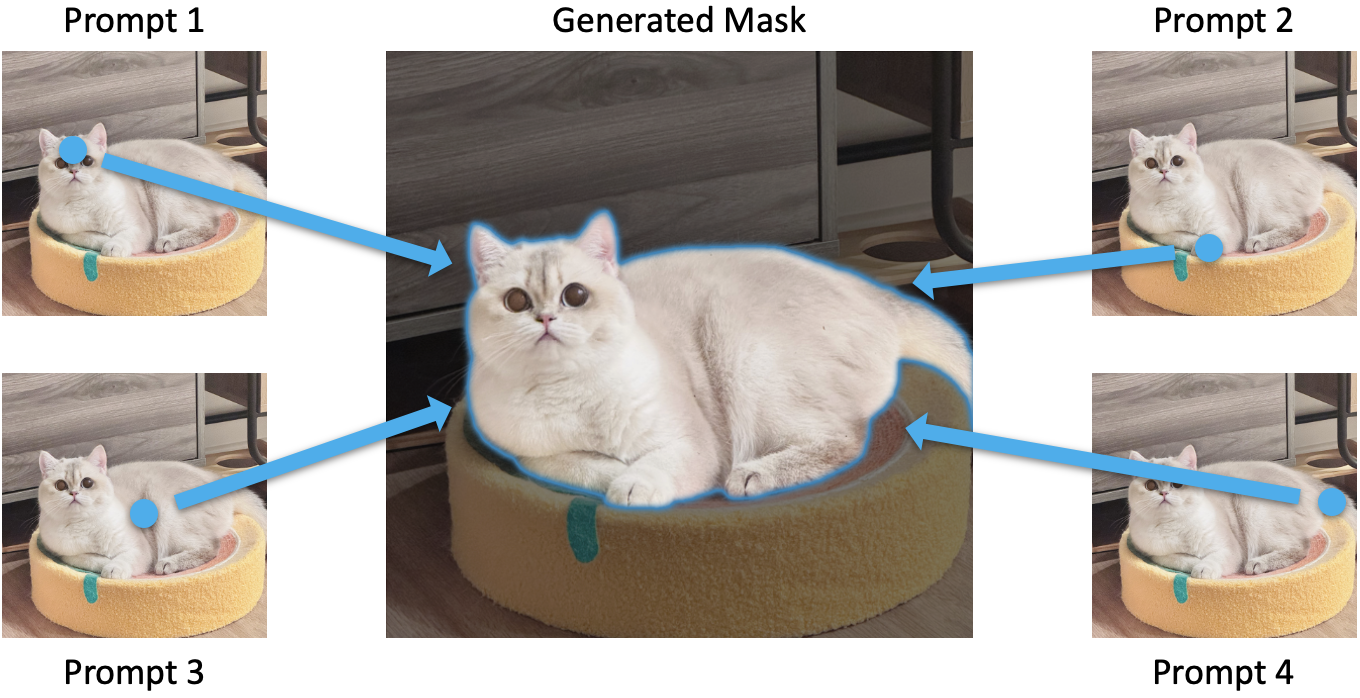}
  \caption{The same mask can be generated by SAM using various prompts in a region.}
  \label{fig:motivation-example}
\end{figure}

To tackle the challenges resulting from long-tailed data shortage and descriptive limitation, image segmentation with visual references (i.e., an annotated reference image that indicates the objects of interest) has become increasingly important. Its ultimate goal is to segment similar objects as indicated in the annotated reference image, regardless of the semantic category of target objects. Recently, various methods~\cite{persam,matcher,vrpsam} have leveraged the exceptional segmentation capabilities of SAM, achieving significant breakthroughs in visual reference segmentation. The main idea of these methods is to generate prompts that direct SAM to predict masks for the target objects. Among these approaches, training-based methods~\cite{vrpsam} that learn prompt embeddings to guide SAM mask generation have achieved state-of-the-art (SOTA) performance. However, the existing training approaches often direct the prompts toward the boundaries of target regions, resulting in instability and reduced robustness. This robustness challenge is specifically caused by the inherent design of SAM --- identical masks can be generated from different prompts (see Figure~\ref{fig:motivation-example}). This oversight poses a significant challenge for the training-based methods, where stability and generalizability are essential for zero-shot capability.

To address this limitation, we introduce a novel visual reference segmentation method, \modelname, to enhance the robustness and zero-shot capability of SAM-based visual reference segmentation. Specifically, inspired by the spirit of variational inference \cite{vi} in statistics, we propose a variational prompt encoder with reparameterization trick to predict a multivariate prompt distribution in high-dimensional space, such that every prompt sampled from this multivariate prompt distribution can effectively guide SAM to generate a high-quality mask for the target object. During inference, the predicted mean of this multivariate prompt distribution will be utilized to generate the predicted mask with the same visual concept as the reference object. Unlike the existing training-based method~\cite{vrpsam}, which does not favor the prompts closer to the center of the target prompt region, our method encourages the mean of the multivariate prompt distribution to be closer to the center of the target prompt region by injecting the noise into the generated prompts and penalizing the Laplacian during training.

To demonstrate the effectiveness of our method, we conducted extensive experiments on Pascal-5$^i$ and COCO-20$^i$  datasets following the same dataset configuration as the SOTA methods~\cite{persam, vrpsam}. The experimental results demonstrate that our approach consistently outperforms the SOTA method on both datasets. In summary, the contributions of this paper are threefold:
\begin{itemize}
  \item We identify a commonly overlooked limitation in the SOTA SAM-based visual reference segmentation approach, where prompts are often generated at boundaries of target regions, leading to instability and reduced robustness.
  \item We propose a probabilistic prompt generation method that leverages variational inference to penalize the prompts that lie in unstable regions, enhancing the robustness of generated prompts.
  \item Our approach consistently outperforms the SOTA SAM-based visual reference segmentation method on the Pascal-5$^i$ and COCO-20$^i$  datasets.
\end{itemize}

\section{Related Work}
\label{sec:related-work}

\subsection{Visual Reference Segmentation}

A visual reference is an annotated reference image that represents the object of interest. Segmenting based on visual reference prompt provides a more intuitive and straightforward way to guide the segmentation of the desired object in the target image, regardless of its semantic category. Unlike text prompts, visual reference prompt bypasses the need for cross-modality alignment between text and image, because it solely relies on visual similarities~\cite{trex2}. This unique strength enhances generalizability in segmenting novel objects that were unseen during training.

With the recent advancements in vision foundation models, several SAM-based methods~\cite{vrpsam,matcher} have achieved significant breakthroughs in visual reference segmentation, by transforming visual references into prompts that SAM can understand. These SAM-based methods can be classified into two categories: training-based approaches~\cite{vrpsam} and training-free approaches~\cite{matcher,persam}. Notably, VRP-SAM~\cite{vrpsam}, a SAM-based training approach, achieves SOTA performance in this task. However, the existing training-based approaches including VRP-SAM fail to consider the robustness of generated prompts. This oversight motivates us to propose a variational prompt encoder, which has been theoretically and empirically demonstrated to generate more robust prompts.

\vspace{-5pt}

\subsection{Variational Inference}

To facilitate the robustness of the prompts generated for SAM, we draw inspiration from the principles of variational inference \cite{vi} in statistics and propose a SAM-based variational prompt encoder to predict a probabilistic prompt distribution based on the visual reference. In the existing literature, the variational inference has been applied across various tasks (e,g., data generation \cite{vae}, metric learning \cite{lin2018deep}, person re-identification \cite{yu2019robust}, and semantic segmentation \cite{xie2023boosting}) with different purposes. Specifically, VAE \cite{vae} and DVML \cite{lin2018deep} adopt variational inference to generate more diverse and discriminative samples by modeling the data variance, while DistributionNet \cite{yu2019robust} and PRCL \cite{xie2023boosting} utilize variational inference to handle the noisy data by estimating the uncertainty of data distribution. Unlike these methods, we leverage variational inference to inject noise into the prompt embeddings with the purpose of penalizing Laplacian during training, such that more robust prompts can be generated to guide SAM. To the best of our knowledge, we are the first SAM-based segmentation method that learns a variational prompt encoder to generate probabilistic prompts, specifically designed to enhance the robustness of SAM mask generation.

\section{Preliminary}
\label{sec:preliminary}

\begin{figure*}[!ht]
  \centering
  \includegraphics[width=0.78\linewidth]{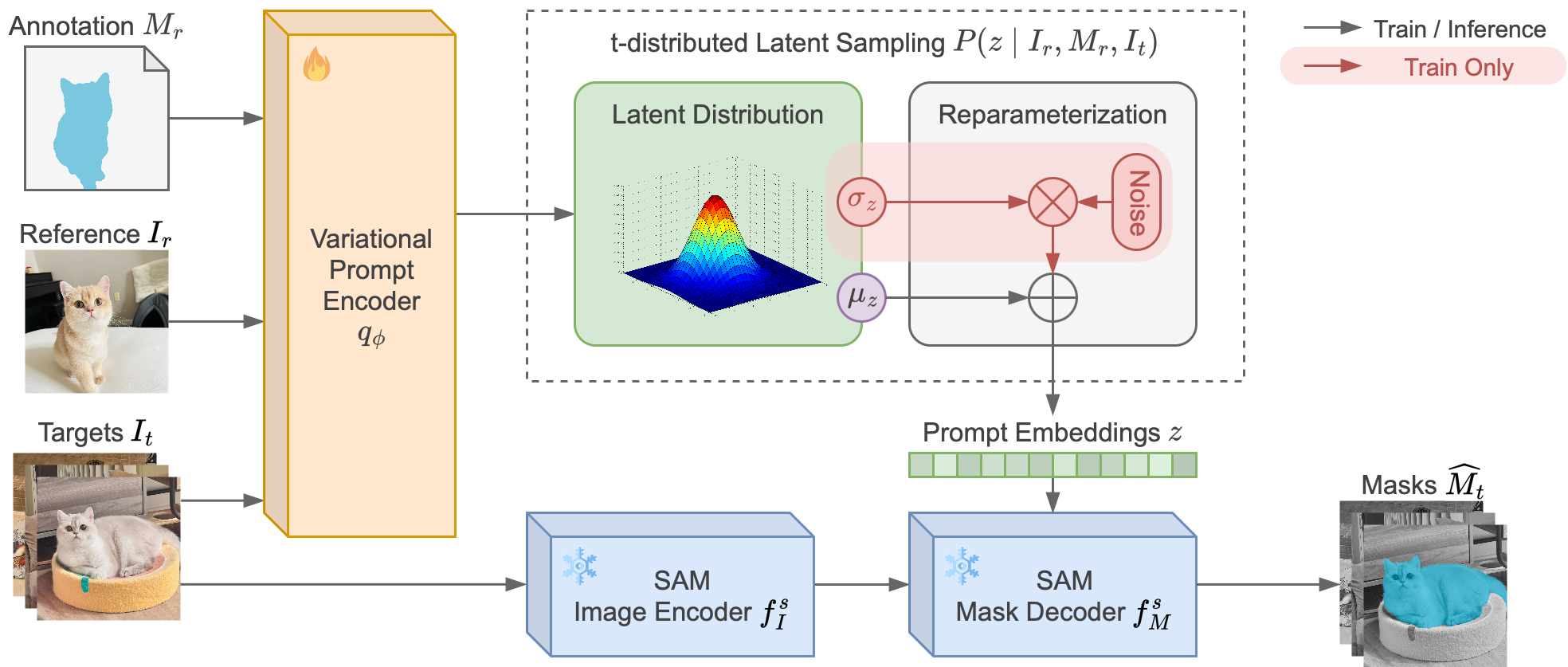}
  \caption{The overview of \modelname, which segments the target images based on the visual references. Given a pre-trained SAM, a variational prompt encoder is trained to predict a multivariate prompt distribution with reparameterization trick. During the inference, the predicted mean prompt is employed to guide SAM in producing robust mask prediction for the target image.}
  \label{fig:overview}
\end{figure*}

\subsection{SAM}

SAM is designed to generate a segmentation mask for a given input image $I$ based on user-specified prompts 
$P$. These prompts can be in various forms, such as points or boxes. Its architecture is composed of three major components: a prompt encoder, an image encoder, and a lightweight mask decoder, denoted by $f_P^{\text{s}}$, $f_I^{\text{s}}$, and $f_M^{\text{s}}$, respectively. Specifically, the image encoder $f_I^{\text{s}}$ extracts features from the input image to produce $F_I^{\text{s}}$, while the prompt encoder $f_P^{\text{s}}$ processes $m$ user-provided prompts to generate prompt embeddings $\bm{z}$. This can be expressed as follows:
\begin{equation}
F_I^{\text{s}} = f_I^{\text{s}}(I), \quad \bm{z} = f_P^{\text{s}}(P),
\end{equation}
where $F_I^{\text{s}} \in \mathbb{R}^{h \times w \times c}$ is the image feature map with resolution $h \times w$ and feature dimension $c$, and $\bm{z} \in \mathbb{R}^{m \times c}$ represents the prompt embeddings. The encoded image features $F_I^{\text{s}}$ and prompt embeddings $\bm{z}$ are then passed into the decoder $f_M^{\text{s}}$ to produce the final mask output, represented as:
\begin{equation}\label{equ:sam-mask-decoder}
\hat{M} = f_M^{\text{s}} \left( F_I^{\text{s}}, \bm{z}\right),
\end{equation}
where $\hat{M}$ represents the final mask predicted by SAM.

\subsection{Existing Training-Based Method Using SAM}
\label{sec:vrpsam}
In the existing literature, VRP-SAM~\cite{vrpsam} is currently the most advanced training-based method, which achieves state-of-the-art performance in this task. Therefore, our method is built on top of VRP-SAM to further enhance the SAM-based visual reference segmentation.

\textbf{Visual Reference Prompt Encoder.} In VRP-SAM, the only trainable module is the visual reference prompt encoder, as it utilizes a pre-trained SAM, along with a pre-trained image encoder (e.g., ResNet-50~\cite{resnet}).  This prompt encoder transforms various annotation formats for reference images (e.g., points, boxes, scribbles, and masks) into high-dimensional prompt embeddings that share the same output space as the SAM prompt encoder. There are two major components in their visual reference prompt encoder: feature augmenter and prompt generator. To be specific, the feature augmenter leverages a semantic-aware image encoder $f_I$ to extract the enhanced image features $F_r^v$ and $F_t^v$ for reference image $I_r$ and target image $I_t$. These enhanced image features, $F_r^v$ and $F_t^v$, 
 capture the object-specific features extracted from visual annotation $M_r$ and a pseudo-mask of target image $M_t^\text{pseudo}$, respectively. Then, the prompt generator $f_P^v$ will output a latent prompt $\bm{z}$ as follows,
\begin{equation}
\bm{z} = f_P^v(F_r^v, F_t^v),\quad \text{for } \bm{z} \in \mathbb{R}^{m \times c}. 
\end{equation}
Lastly, given the generated prompt embeddings $\bm{z}$, a mask prediction $\hat{M}_t$ is generated as in \autoref{equ:sam-mask-decoder}. The predicted mask $\hat{M}_t$ is expected to encapsulate a visual concept similar to that of the visual reference ($I_r$, $M_r$). Notably, with $m$ set to 50 as the default value, the prompt encoder $f_P^v$ predicts 50 prompt embeddings for each target image, enabling a more comprehensive representation of the visual characteristics of the reference objects.

\textbf{Loss Function.} 
To supervise the learning of its prompt encoder $f^v_P$, Binary Cross-Entropy loss and Dice loss are computed between the predicted mask $\hat{M}_t$ and ground-truth mask $M_t$ as below,
\begin{equation}\label{equ:vrpsam-loss}
\mathcal{L} = \mathcal{L}_{\text{BCE}}(\hat{M}_t, M_t) + \mathcal{L}_{\text{Dice}}(\hat{M}_t, M_t).
\end{equation}
In essence, VRP-SAM focuses solely on mask-level differences, while overlooking the potential for further optimizing the prompt encoder to generate more robust prompts.

\section{\modelname}
\label{sec:method}

In this paper, our high-level objective is to automatically generate robust prompts to guide SAM in producing high-quality segmentation masks containing the same visual concepts as the visual reference. To this end, we first identify a unique robustness challenge for the SAM-based segmentation method (see \autoref{sec:robust-challenge}). To address this challenge, we propose a variational prompt encoder in \autoref{sec:vpe} that transforms the visual reference into a multivariate prompt distribution, such that the predicted mean prompt can be employed to generate high-quality and robust masks during inference. The model training and inference procedures are described in \autoref{sec:train-inference}.

\subsection{Robustness Challenge}
\label{sec:robust-challenge}
Robust prompts are crucial for the SAM model to yield stable and precise final mask predictions. Yet, the importance of prompt robustness has been largely overlooked in the existing literature, as pointed out in \autoref{sec:vrpsam}. In this section, we pinpoint this critical challenge in existing SAM-based segmentation methods: the failure to account for the robustness of generated prompts.

In practice, there exists a region of prompt embeddings in the high-dimensional space in which every prompt can lead SAM to produce acceptable segmentation masks (see \autoref{fig:motivation-example}). This region is referred to as the target prompt region \(\mathcal{R}_{I_r,M_r,I_t}\). However, the stability and robustness of different prompts within this region can vary. For less robust prompts, even a small perturbation can lead to significant changes in the masks produced by SAM. This instability is more prevalent when the learned prompts lie near the boundary of the target prompt region. Such unstable prompts in the boundary areas are more likely to be generated if a learnable prompt generator is not explicitly guided to produce prompts toward the center of \(\mathcal{R}_{I_r,M_r,I_t}\) (see Appendix~\ref{sec:vrp-drawback} for detailed analysis). Theoretically speaking, the loss landscape in the boundary areas typically exhibits sharp gradient variations due to high curvature (quantified by the Hessian \(\nabla^2 \mathcal{L}(z)\)), given a loss function $\mathcal{L}$ that solely considers the mask-level differences (e.g., \autoref{equ:vrpsam-loss}).

Ideally, the prompt embedding should reside in a flat region of the loss landscape with low curvature, which is particularly important when generalizing to objects with unseen semantic categories. One straightforward way to enforce this would be to directly regularize the curvature by penalizing the Hessian \(\nabla^2 \mathcal{L}(z)\). However, incorporating such a curvature-based regularization term directly into the loss function is challenging due to the limitation of the automatic differentiation-based deep learning framework \cite{auto-diff}, because it requires computing second-order derivatives. Unfortunately, no existing approaches including VRP-SAM attempt to encourage prompt embeddings to lie in flatter regions of the loss landscape. This issue is a challenging but important gap that we aim to bridge in this work.

\begin{figure}[tb]
  \centering
  \includegraphics[width=0.78\linewidth]{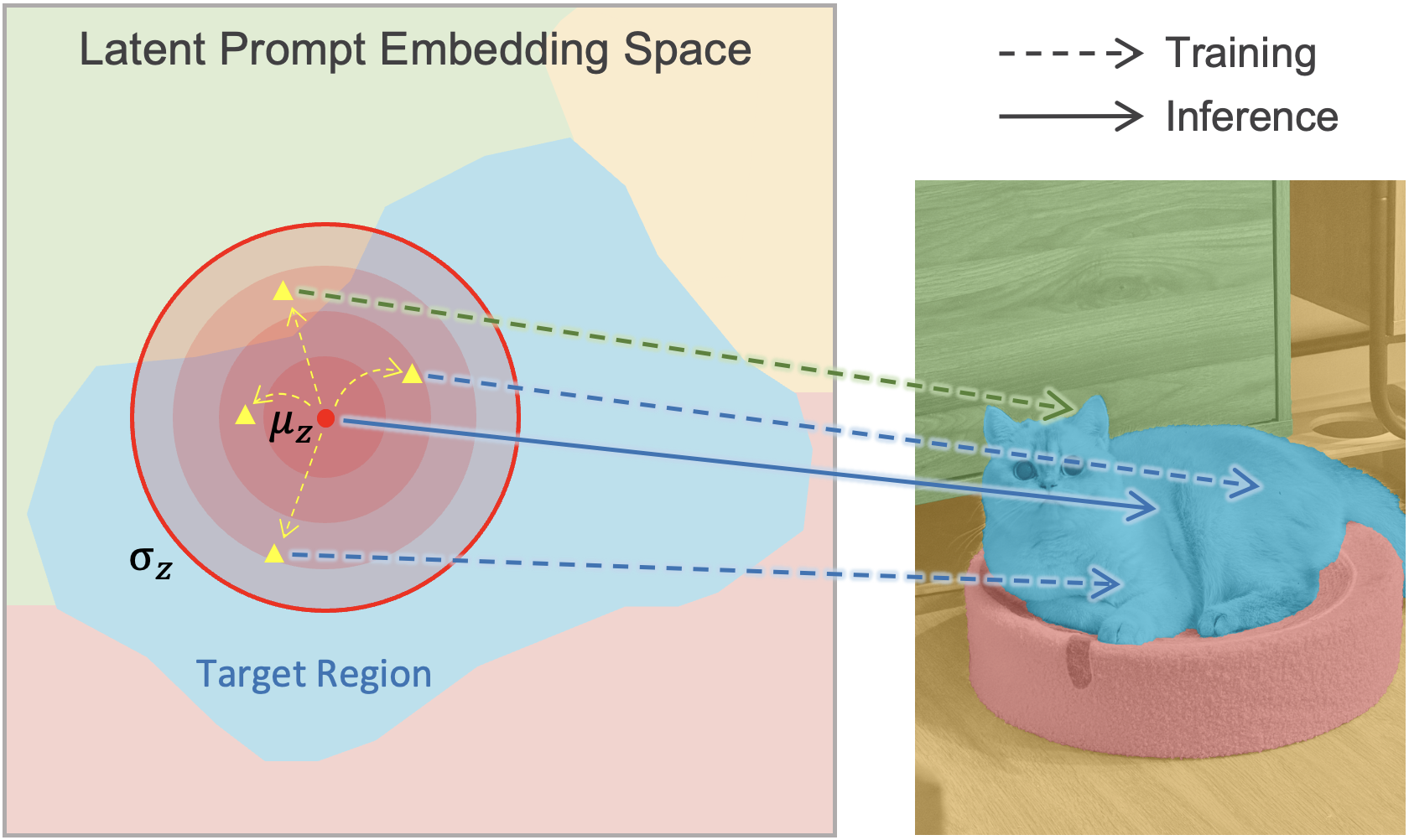}
  \caption{Intuitive illustration of the high-level idea behind the proposed variational prompt encoder. The dashed arrow shows the sampling and prompting procedure during training, while the solid arrow shows the prompting strategy during inference. Note that our generated prompts can be any type of prompt, while this illustration shows a positive point prompt as an example.}
  \label{fig:vpe-intuition}
\end{figure}

\subsection{Variational Prompt Encoder}
\label{sec:vpe}

To improve the robustness of generated prompts, we propose a simple but effective method to learn a more robust prompt encoder that favors prompts in the flatter regions of the loss landscape without explicitly incorporating a second-order regularization term. To be specific, we introduce a variational prompt encoder, denoted as \(q_{\phi}(\bm{z}| I_r, M_r, I_t)\) and parameterized by \(\phi\), to approximate the true multivariate prompt distribution \(P(\bm{z}| I_r, M_r, I_t)\). The objective is to maximize the likelihood that a sampled prompt embedding \(\bm{z} \sim q_{\phi}(\bm{z}| I_r, M_r, I_t)\) falls within the target prompt region \(\mathcal{R}_{I_r, M_r, I_t}\). As illustrated in \autoref{fig:vpe-intuition}, a straightforward intuition behind this framework is to optimize the robustness of expected prompt embedding \(\bm{\hat{\mu}}_{\bm{z}}\) by inducing a margin between \(\bm{\hat{\mu}}_{\bm{z}}\) and the boundary of \(\mathcal{R}_{I_r, M_r, I_t}\) leveraging the standard deviation \(\bm{\hat{\sigma}}_{\bm{z}}\) of the variational prompt distribution \(q_{\phi}(\bm{z}| I_r, M_r, I_t)\). This makes the mean prompt $\bm{\hat{\mu}}_{\bm{z}}$ less likely to fall outside the target region under small perturbations. Furthermore, based on the theoretical analysis presented in \autoref{sec:equ-noise-lap} and \autoref{sec:prosam-robust} in Appendix, we prove that performing variational optimization leads to an implicit penalty on the curvature of the loss function, thereby encouraging the optimization to favor flatter regions in the loss landscape, which is usually closer to the center of \(\mathcal{R}_{I_r,M_r,I_t}\). Empirically, our verification study also showcases that our framework effectively pushes prompt embeddings toward the target region center (see \autoref{sec:verify-study}.) Therefore, based on both the theoretical analysis and the empirical results, the proposed variational prompt encoder naturally addresses the challenge we presented in \autoref{sec:robust-challenge}.

A straightforward instantiation of $P(\bm{z}|I_r,M_r,I_t)$ is assuming $\bm{z}$ follows a conditional multivariate Gaussian distribution. However, considering our motivation is to push the mean prompt $\bm{\hat\mu}_{\bm{z}}$ toward the center of the target prompt embedding region, a more heavy-tailed t-distribution would be a better choice due to its statistical properties of having more chance to sample outliers~\cite{conditional-t}. In other words, when $\bm{\hat\mu}_{\bm{z}}$ is close to the boundary of $\mathcal{R}_{I_r,M_r,I_t}$, it is more likely to sample a prompt falling outside the target prompt embedding region (green region in \autoref{fig:vpe-intuition}) if it follows heavy-tailed t-distribution. Given that the outlier prompt is more likely to result in low-quality mask predictions, the force to push the mean prompt toward the center of $\mathcal{R}_{I_r,M_r,I_t}$ will be greater with larger gradients. Theoretically, we also show in \autoref{sec:t-advantage} that t-distribution results in a larger 4th-order curvature penalty that forces an extra push toward flatter and more stable regions. Inspired by Kim 
 et al.~\cite{t3vae}, we formulate our variational prompt distribution $q_{\phi}(\bm{z}|I_r,M_r,I_t)$ with the heavy-tailed property and diagonal covariance as follows,
\begin{equation}\label{equ:prompt-dist}
q_{\phi}(\bm{z}|I_r,M_r,I_t) = t \bigg(\bm{z}\bigg|\bm{\mu}_{\bm{z}},\frac{\text{diag}\left(\bm{\sigma}_{\bm{z}}^{2}\right)}{1+\nu^{-1}n},\nu+n \bigg)
\end{equation}
where $\nu$ is a hyper-parameter to control the degree of heavy-tailless and $n$ is the dimensionality of $\bm{z}$.

To learn a variational prompt distribution via back-propagation, the sampling function for sampling $\bm{z}$ from $q_{\phi}(\bm{z}|I_r,M_r,I_t)$ must be differentiable. Thus, the reparameterization trick \cite{vae-st} is employed to approximate the sampling process of $\bm{z}$ with a differentiable function. With two independent random variables $\bm\epsilon \overset{\text{iid}}{\sim} \mathcal{N}(0,I)$ and $\bm\delta \overset{\text{iid}}{\sim} \chi^2(\nu + n)$ as the source of randomness, the sampling function $g_{\bm{z}}$ for drawing $\bm{z}$ from $q_{\phi}(\bm{z}|I_r,M_r,I_t)$ can be formulated as,
\begin{align}
\bm{z} &= g_{\bm{z}}(\bm{\hat\mu}_{\bm{z}}, \bm{\hat\sigma}_{\bm{z}}, \bm\delta, \bm\epsilon) \\
&= \bm{\hat\mu}_{\bm{z}} + \frac{1}{\sqrt{\bm\delta} / (\nu + n)} \frac{\bm{\hat\sigma}_{\bm{z}}}{\sqrt{1 + \nu^{-1} n}} \odot \bm\epsilon. \label{equ:reparam-trick}
\end{align}
This formulation enables us to sample $\bm{z} \sim q_{\phi}(\bm{z}|I_r,M_r,I_t)$ as a differentiable function of $\bm{\hat\mu}_{z}$ and $\bm{\hat\sigma}_{\bm{z}}$, predicted by the variational prompt encoder.

\subsection{Training and Inference}
\label{sec:train-inference}

\textbf{Training.} During training, our goal is to minimize the expected loss w.r.t. the prompt embedding distribution,
\begin{align}\label{equ:objective}
    &\min_{\mathbf{\phi}} \mathbb{E}_{\bm{z} \sim q_{\phi}(\bm{z}|I_r,M_r,I_t)} \left[
        \mathcal{L}\left(f_M^{s}(\bm{z}, F_{I_t}^s),M_{t}\right)
    \right] \\
    =&
    \min_{\mathbf{\phi}} \int_{\mathbb{R}^n} q_{\phi}(\bm{z}|I_r,M_r,I_t)
        \mathcal{L}\left(f_M^{s}(\bm{z}, F_{I_t}^s),M_{t}\right)  d\bm{z},
\end{align}
where $F_{I_t}^s$ = $f_I^{s}(I_t)$. However, this integral is intractable. Therefore, we employ the Monte Carlo method to approximate it by minimizing the expected loss with K samples,
\begin{align}
    & \min_{\mathbf{\phi}} \mathbb{E}_{\bm{z} \sim q_{\phi}(\bm{z}|I_r,M_r,I_t)} \left[
        \mathcal{L}\left(f_M^{s}(\bm{z}, F_{I_t}^s),M_{t}\right)
    \right] \notag \\
    =& \min_{\mathbf{\phi}} \mathbb{E}_{\bm\epsilon \overset{\text{iid}}{\sim} \mathcal{N}(0,I), \bm\delta \overset{\text{iid}}{\sim} \chi^2(\nu + n)} \left[
        \mathcal{L}\left(f_M^{s}(\bm{z}, F_{I_t}^s),M_{t}\right)
    \right] \notag \\
    \approx&\ \resizebox{0.83\linewidth}{!}{$
    \min_{\mathbf{\phi}} \frac{1}{K}\sum^K_{k=1} \mathcal{L}\big(f_M^{s}(g_{\bm{z}}(\bm{\hat\mu}_{\bm{z}}, \bm{\hat\sigma}_{\bm{z}}, \bm\delta_k, \bm\epsilon_k),F_{I_t}^s),M_{t}\big)
    $}.
\end{align}
Note that the loss function can theoretically be any mask-level loss that evaluates the deviation between the predicted mask $\hat{M_{t}}$ and $M_{t}$. Following VRP-SAM~\cite{vrpsam}, we adopt the BCE loss and Dice loss to enforce both pixel-wise accuracy and the degrees of overlap (see \autoref{equ:vrpsam-loss}). Regarding the model architecture of the variational prompt encoder, our approach theoretically can be applicable to any model architecture. In this paper, to ensure a fair comparison with the existing training-based method, we employ an identical model architecture as VRP-SAM prompt encoder (as described in \autoref{sec:vrpsam}), except adding two linear layers at the end to predict $\bm{\hat\mu}_{\bm{z}}$ and $\bm{\hat\sigma}_{\bm{z}}$ respectively.

\textbf{Inference.} During inference, only the predicted mean prompt $\bm{\hat\mu}_{\bm{z}}$ is used to prompt the SAM mask decoder $f_M^{s}$ to generate a robust mask prediction $\hat{M_{t}}$. This inference strategy facilitates the robustness of generated prompts for novel objects, leveraging the margin between $\bm{\hat\mu}_{\bm{z}}$ and the boundary of target prompt region $\mathcal{R}_{I_r,M_r,I_t}$. More importantly, since we rely solely on the predicted mean prompt for inference, our inference speed and memory usage are on par with that of non-probabilistic methods such as VRP-SAM~\cite{vrpsam}.

\section{Experiments}
\label{sec:result}

\subsection{Experimental Setup}
\label{sec:experiment-setup}

\begin{table}[ht!]
\renewcommand\arraystretch{0.85}
\centering
\caption{Quantitative comparison with SOTA visual reference segmentation methods based on mIoU. $^\dagger$ represents the method is based on SAM. For the models trained by in-domain datasets, their results are colored in \textcolor{gray}{gray}. The colors \best{green} and \sbest{blue} indicate the best and second-best results, respectively, among all methods that were not trained with in-domain datasets.}
\setlength{\tabcolsep}{0.3\tabcolsep}
\tiny
\resizebox{\linewidth}{!}{%
\begin{tabular}{l l ccccc c}

\toprule

\multirow{2}{*}{Data} & \multirow{2}{*}{Methods}  & \multirow{2}{*}{\makecell[c]{Label\\Type}} &  \multirow{2}{*}{F-0} & \multirow{2}{*}{F-1} & \multirow{2}{*}{F-2} & \multirow{2}{*}{F-3} & \multirow{2}{*}{Means} \\
\\
\midrule

\multirow{13}{*}{COCO-20$^i$} &  \textcolor{gray}{Painter\cite{painter}} & \multirow{5}{*}{mask} & \textcolor{gray}{31.2} & \textcolor{gray}{35.3} & \textcolor{gray}{33.5} & \textcolor{gray}{32.4} & \textcolor{gray}{33.1} \\
& \textcolor{gray}{SegGPT\cite{seggpt}} & & \textcolor{gray}{56.3} & \textcolor{gray}{57.4} & \textcolor{gray}{58.9} & \textcolor{gray}{51.7} & \textcolor{gray}{56.1} \\
& PerSAM$^\dagger$\cite{persam} & & 23.1 & 23.6 & 22.0 & 23.4 & 23.0 \\
& PerSAM-F$^\dagger$\cite{persam} & & 22.3 & 24.0 & 23.4 & 24.1 & 23.5 \\
& Matcher$^\dagger$\cite{matcher} & & \best{52.7} & 53.5 & 52.6 & \sbest{52.1} & 52.7 \\

\cmidrule(lr){2-8}
& \multirow{4}{*}{VRP-SAM$^\dagger$\cite{vrpsam}}   & point & 32.03 & 39.36 & 46.44 & 40.52 & 39.85 \\
&  & scribble & 44.83 & 48.22 & 51.61 & 47.66 & 48.08 \\
&  & box & 44.63 & 49.2 & 56.56 & 49.34 & 49.93 \\
& & mask & 47.02 & \sbest{54.24} & \sbest{59.91} & 51.95 & \sbest{53.28} \\
\cmidrule(lr){2-8}
& \cc{} & \cc{ point} &  \cc{33.32} &  \cc{40.23} &  \cc{47.82} & \cc{41.2} & \cc{40.64} \\
& \cc{} &\cc{ scribble} & \cc{47.37} & \cc{48.97} & \cc{53.44} & \cc{48.39} & \cc{49.54}\\
& \cc{} & \cc{box} & \cc{45.39} & \cc{50.01} & \cc{57.92} & \cc{50.41} & \cc{50.93}\\
&  \cc{\multirow{-4}{*}{\modelname$^\dagger$}} &\cc{ mask} & \cc{\sbest{48.74}} & \cc{\best{55.55}} & \cc{\best{60.72}} & \cc{\best{53.49}} & \cc{\best{54.63}} \\

\midrule
\midrule

\multirow{8}{*}{PASCAL-5$^i$} & \multirow{4}{*}{VRP-SAM$^\dagger$\cite{vrpsam}}   & point & 63.69 & 70.95 & 63.22 & 54.53 & 63.35 \\
&  & scribble & 70.04 & 74.67 & 65.93 & 59.12 & 67.44 \\
& & box & 71.3 & 75.98 & 65.95 & 61.27 & 68.75 \\
& & mask & \sbest{74.01} & 76.77 & \sbest{69.46} & \sbest{64.34} & \sbest{71.14} \\
\cmidrule(lr){2-8}
&\cc{} & \cc{point} & \cc{64.71} & \cc{72.11} & \cc{63.89} & \cc{55.64} & \cc{64.08}\\
& \cc{} & \cc{scribble} & \cc{71.16} & \cc{75.69} & \cc{66.31} & \cc{60.93} & \cc{68.52}\\
& \cc{} & \cc{box} &\cc{ 72.38} & \cc{\sbest{76.81}} & \cc{67.07} &\cc{ 62.76} & \cc{69.76}\\
&  \cc{\multirow{-4}{*}{\modelname$^\dagger$}}   & \cc{mask} & \cc{\best{75.26}} &\cc{ \best{77.57}} &\cc{ \best{70.09}} & \cc{\best{65.22}} & \cc{\best{72.04}} \\
\bottomrule
\end{tabular}%
}
\label{tab:fm-result}
\end{table}

\begin{table*}[ht!]
\renewcommand\arraystretch{0.8}
\centering
\caption{Quantitative comparison with one-shot segmentation methods based on mIOU. The \best{green} and \sbest{blue} colors indicate the best and second-best results, respectively.}
\setlength{\tabcolsep}{\tabcolsep}

\resizebox{0.8\linewidth}{!}{%
\begin{tabular}{c c c ccccc ccccc}
\toprule
\multirow{2}{*}{Methods} & \multirow{2}{*}{\makecell[c]{Image \\ Encoder}} & \multirow{2}{*}{\makecell[c]{Learnable\\ Params}} & \multicolumn{5}{c}{COCO-20$^i$} & \multicolumn{5}{c}{PASCAL-5$^i$} \\
\cmidrule(lr){4-8} \cmidrule(lr){9-13}
 & & & F-0 & F-1 & F-2 & F-3 & Mean & F-0 & F-1 & F-2 & F-3 & Mean \\

\midrule
PFENet \cite{pfenet} & \multirow{10}{*}{ResNet-50} & 10.4M & 36.5 & 38.6 & 34.5 & 33.8 & 35.8 & 61.7 & 69.5 & 55.4 & 56.3 & 60.8 \\
HSNet \cite{hsnet} &  & 2.6M & 36.3 & 43.1 & 38.7 & 38.7 & 39.2 & 64.3 & 70.7 & 60.3 & 60.5 & 64.0 \\
CyCTR \cite{cyctr}& & 15.4M & 38.9 & 43.0 & 39.6 & 39.8 & 40.3 & 65.7 & 71.0 & 59.5 & 59.7 & 64.0 \\
SSP \cite{ssp} &  & 8.7M & 35.5 & 39.6 & 37.9 & 36.7 & 37.4 & 60.5 & 67.8 & 66.4 & 51.0 & 61.4 \\
NTRENet \cite{ntrenet} &  & 19.9M & 36.8 & 42.6 & 39.9 & 37.9 & 39.3 & 65.4 & 72.3 & 59.4 & 59.8 & 64.2 \\
DPCN \cite{dpcn} &  & - & 42.0 & 47.0 & 43.3 & 39.7 & 43.0 & 65.7 & 71.6 & 69.1 & 60.6 & 66.7 \\
VAT \cite{vat} &  & 3.2M & 39.0 & 43.8 & 42.6 & 39.7 & 41.3 & 67.6 & 72.0 & 62.3 & 60.1 & 65.5 \\
BAM \cite{bam} &  & 4.9M & 39.4 & 49.9 & 46.2 & 45.2 & 45.2 & 69.0 & 73.6 & 67.6 & 61.1 & 67.8 \\
HDMNet \cite{hdmnet} &  & 4.2M & 43.8 & \sbest{55.3} & \sbest{51.6} & \sbest{49.4} & 50.0 & 71.0 & \sbest{75.4} & \sbest{68.9} & 62.1 & \sbest{69.4} \\

\ourrow \modelname &  & 1.73M & \sbest{48.74} & \best{55.55} & \best{60.72} & \best{53.49} & \best{54.63} & \best{75.26} & \best{77.57} & \best{70.09} & \sbest{65.22} & \best{72.04} \\

\midrule
DCAMA~\cite{dcama} & Swin-B & 47.7M & \best{49.5} & 52.7 & 52.8& 48.7& \sbest{50.9} & \sbest{72.2} & 73.8 & 64.3 & \best{67.1} & 69.3\\

\bottomrule
\end{tabular}%
}
\label{tab:fs-result}
\end{table*}

\textbf{Datasets.} To evaluate the effectiveness and generalizability of \modelname, we conducted comprehensive experiments on Pascal-5$^i$~\cite{pascal} and COCO-20$^i$ ~\cite{coco} under the same few-shot setting as the existing visual reference segmentation methods~\cite{vrpsam,lang2022learning,zhang2021few}. In this setting, these two datasets are divided into 4 folds. In each fold, Pascal-5$^i$ includes 15 base classes for training and 5 novel classes for testing, while COCO-20$^i$  has 60 base classes for training and 20 novel classes for testing. Therefore, the robustness and generalizability of each visual reference segmentation method can be fully assessed under this setting. Following VRP-SAM~\cite{vrpsam}, 1,000 pairs of visual reference and target images are randomly selected to evaluate our testing performance for each fold.

\textbf{Implementation Details.} To ensure a strictly fair comparison with VRP-SAM, the SOTA method in the visual reference segmentation, we ensure that all the experimental settings (e.g., random seed, LR scheduler, optimizer) and hyper-parameters (e.g., number of layers, prompt embedding dimensions, number of prompts) are identical to VRP-SAM. Specifically, the model architecture of the variational prompt encoder is similar to VRP-SAM (\textit{see \hyperref[sec:model-archi]{Section~\ref{sec:model-archi}} in Appendix for more details}), except two linear layers have been added at the end of the prompt encoder to predict the mean and log standard deviation of the multivariate prompt distribution. In other words, our method is easy to implement with only a few lines of code, but can effectively boost the robustness of generated prompts. During training, the number of Monte Carlo samples K has been set to 10, with the degrees of freedom $\nu$ of 5. Also, we employ the AdamW~\cite{adamw} optimizer with weight decay of 1e-6 and the cosine annealing learning rate scheduler with warm restart~\cite{lr-scheduler} after 15 epochs, and the initial learning rate of 1e-4. The model is trained with 100 epochs with a fixed random seed of 321. For the choice of image encoder utilized by the variational prompt encoder, ResNet-50~\cite{resnet} is adopted following VRP-SAM. In addition, our variational prompt encoder predicts 50 multivariate prompt distributions per target image and only uses 50 mean prompts during inference, which guarantees the same number of prompts have been used in the experimental study for both VRP-SAM and \modelname. In terms of the visual annotation types, we follow the same procedure as VRP-SAM~\cite{vrpsam} and SEEM~\cite{seem} to automatically generate points, scribbles, and boxes based on the mask annotations. Lastly, the mean intersection over union (mIOU) is adopted to evaluate our segmentation performance across all datasets. All experiments on COCO-20$^i$ were conducted on 4 RTX 4090 GPUs with a batch size of 2 per GPU, whereas the experiments on Pascal-5$^i$ were conducted on 1 H100 GPU with a batch size of 2.  On a single H100 GPU with batch size 2, ProSAM takes 0.19s per training batch and 0.12s per inference batch.

\subsection{Quantitative Evaluation}
\label{sec:quant-result}

To assess the effectiveness of \modelname, we compare our method against the existing visual reference segmentation methods via mIOU. Specifically, for both Pascal-5$^i$ and COCO-20$^i$, \modelname\ is trained and tested for each fold separately to evaluate the models on the \textit{unseen classes only}. Unfortunately, we are unable to reproduce the VRP-SAM results they reported in their paper, possibly due to the different hardware we had. Therefore, in order to ensure the fairness of our quantitative comparison against VRP-SAM, we reported the experimental results of both VRP-SAM and our method under completely identical experimental settings as mentioned in \autoref{sec:experiment-setup}. The only difference is the hyper-parameters introduced by learning multivariate prompt t-distributions (i.e., Monte Carlo Samples K and degrees of freedom $\nu$), because those are not applicable to VRP-SAM. For qualitative evaluations, please refer to \hyperref[sec:quail-result]{Section~\ref{sec:quail-result}} in the Appendix.

\textbf{Quantitative Comparison with FM-based Methods.} Powered by the recent advancement of vision foundation models, several visual reference segmentation methods achieved a performance breakthrough~\cite{matcher,seggpt,persam,vrpsam,painter}. According to \autoref{tab:fm-result}, our method with mask annotation achieves the best performance among all SAM-based methods, and surprisingly obtains a comparable result as SegGPT that is trained and tested on the same set of classes within each fold. Compared with Matcher~\cite{matcher}, which employs a larger image encoder DINOv2 ViT-L, we can still achieve a superior performance using a more lightweight, non-ViT backbone, ResNet-50. Compared with VRP-SAM, our variational prompt encoder enables us to consistently outperform VRP-SAM on both datasets, no matter which type of visual reference has been used. Also, the comparison over the confusion matrix against VRP-SAM (see \autoref{sec:conf-matrix} in Appendix) showcases that \modelname\ improves the balanced performance by reducing both the false negatives and false positives. 

\textbf{Quantitative Comparison with Few-Shot Methods.} To comprehensively evaluate our method, we also compare \modelname\ with the SOTA few-shot segmentation methods. As shown in \autoref{tab:fs-result}, we achieve SOTA performance on both datasets with the least number of learnable parameters. Note that all the experimental results are computed on \textit{novel classes only}, which demonstrate the robustness and generalizability of our variational prompt encoder. Especially compared with DCAMA~\cite{dcama}, which has 47.7M learnable parameters, we can still outperform it with only 1.73M learnable parameters.

\textbf{Generalizability Study Under Domain Shift.}  The generalization capability of \modelname\ under domain shift is critical, as it highlights our performance in scenarios where there is a substantial difference between the domains of the training data and testing data. Following the previous works~\cite{vrpsam,hsnet,pfenet}, we conducted the generalization study under the domain shift from COCO-20$^i$ to PASCAL-5$^i$. Specifically, the models are trained on COCO-20$^i$ and exclusively tested on novel classes from PASCAL-5$^i$ that were not included in COCO-20$^i$ during training. As shown in \autoref{tab:coco2pascal}, \modelname\ with ResNet-50 is able to outperform all other methods, even though their performance on mIOU is already sufficiently high. This result demonstrates the extraordinary generalization capability of \modelname\ under the significant domain difference between training and testing.

\begin{table}[ht]
\renewcommand\arraystretch{0.7}
\centering
\caption{The generalization evaluation under the domain shift from COCO-20$^i$ to PASCAL-5$^i$.  For all methods, mask annotations are employed as the visual reference. The average mIOU across 4 folds is reported as the evaluation metric.}
\label{tab:domain_shift}

\setlength{\tabcolsep}{1.5\tabcolsep}
\resizebox{0.65\linewidth}{!}{%
\begin{tabular}{cccc}

\toprule
Methods & Image Encoder & Mean mIOU \\ 
\midrule
\ourrow \modelname & ResNet-50 & \best{77.65}\\

\midrule
VRP-SAM~\cite{vrpsam}
& \multirow{5}{*}{ResNet-50}
& \sbest{76.44}\\
RPMM~\cite{rpmm}   &   & 49.6 \\ 
PFENet~\cite{pfenet} &  & 61.1 \\ 
RePRI~\cite{repri}   &  & 63.2 \\ 
VAT-HM~\cite{vat-hm}  &  & 65.1 \\ 
\midrule
HSNet~\cite{hsnet}   & \multirow{2}{*}{ResNet-101} & 64.1 \\ 
DGPNet~\cite{dgpnet} &  & 70.1 \\ 
\midrule
FP-Trans~\cite{fp-trans} & DeiT-B/16 & 69.7 \\  
\bottomrule
\end{tabular}
}
\label{tab:coco2pascal}
\vspace{-7pt}
\end{table}

\subsection{Verification Study}
\label{sec:verify-study}

As described in \autoref{sec:vpe}, the primary motivation behind the proposed variational prompt encoder is to generate more robust prompts that lie away from the boundaries of the target prompt region $\mathcal{R}_{I_r,M_r,I_t}$. To empirically validate this, we conducted two complementary studies to assess the robustness of prompts generated by \modelname. Due to the page limit, additional verification studies are included in Appendix \hyperref[sec:more-verify-study]{Section~\ref{sec:more-verify-study}}.

\begin{figure}[h]
  \centering
  \includegraphics[width=0.7\linewidth]{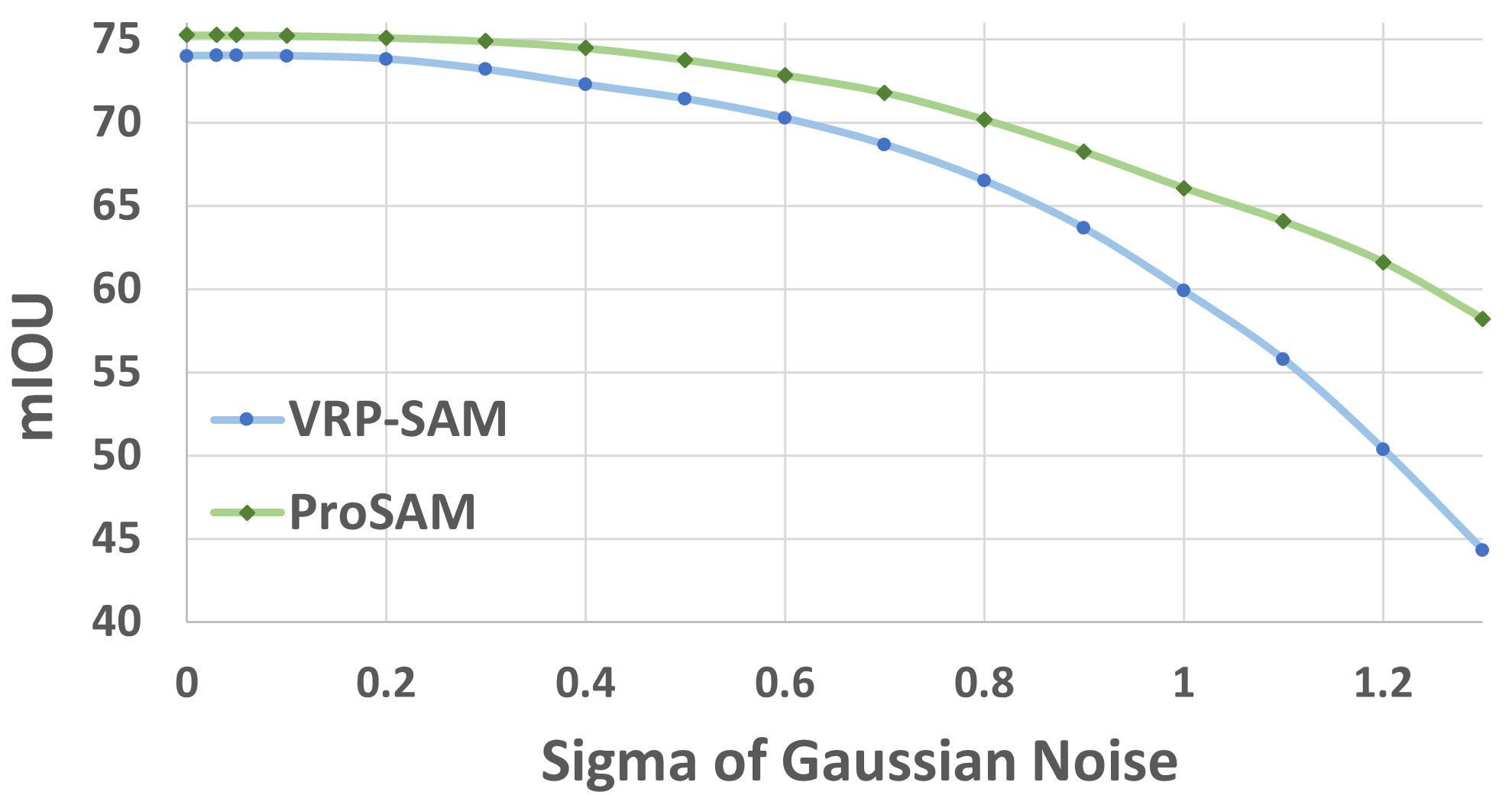}
  \caption{Gaussian noise perturbation on \modelname\ prompts and VRP-SAM prompts in the latent space. Both \modelname\ model and the VRP-SAM model are trained on PASCAL-5$^i$ F-0, and the noise perturbation is injected during inference.}
  \label{fig:robustness}
\end{figure}

\textbf{Robustness to Noise Perturbation}. If our generated prompts truly stay away from the boundaries of the target prompt region, then a small perturbation applied in the latent prompt embedding space should result in minimal degradation in mask quality. To evaluate this, we injected Gaussian noise into the prompt embeddings of both ProSAM and VRP-SAM during inference and analyzed their robustness. As shown in \autoref{fig:robustness}, when the standard deviation of Gaussian noise is set to 1.2, the mIoU of VRP-SAM degrades by approximately 40\%, while \modelname\ only shows a 20\% drop. Overall, VRP-SAM exhibits a much greater sensitivity to noise, suggesting that its prompts tend to lie near the boundaries of target prompt regions, whereas \modelname\ prompts remain significantly more stable under perturbations, providing evidence that they are located closer to the center of the target prompt region.

\textbf{Proximity to Center Prompts}. We further assess prompt robustness by measuring the similarity between the predicted prompts and center point prompts in the ground-truth masks. Since the center prompt of the target prompt regions is unobservable, we approximate it using the latent prompt embedding of the center point prompt in the ground-truth masks because those are the most robust prompts we can obtain. Specifically, we identified the center pixels in the ground-truth masks and fed them into the SAM prompt encoder to get their corresponding prompt embeddings. Then, for each fold in PASCAL-5$^i$, we computed the cosine similarity between the predicted prompts and the embeddings of the center point prompts. As shown in \autoref{tab:cosine_similarity}, \modelname\ prompts consistently exhibit higher cosine similarity to the center point prompts than VRP-SAM prompts. This observation further supports our hypothesis that \modelname\ generates prompts more aligned with the central and robust regions of the target prompt region. Additionally, we evaluated the statistical relationship between cosine similarity and segmentation quality by computing the Pearson correlation coefficient between cosine similarity and mIoU. A positive correlation coefficient of 0.365 suggests that prompts closer to the center point prompt are more likely to yield high-quality segmentation results.

\begin{table}[h]
  \caption{The average cosine similarity between predicted prompts and the embeddings of center point prompts in ground-truth mask.}
\setlength{\tabcolsep}{2\tabcolsep}
  \centering
  \resizebox{0.6\linewidth}{!}{
    \begin{tabular}{c c c}  
      \toprule
      \multirow{2}{*}{\makecell[c]{Cosine Similarity \\ to Center Prompts}} & \multicolumn{2}{c}{Methods} \\
      \cmidrule{2-3}
      & VRP-SAM & ProSAM \\
      \midrule
      F-0 & 0.0072 & \best{0.0122} \\
      F-1 & 0.0127 & \best{0.0291} \\
      F-2 & -0.0040 & \best{0.0341} \\
      F-3 & 0.0106 & \best{0.0420} \\
      \bottomrule
    \end{tabular}
  }
\vspace{-7pt}
  \label{tab:cosine_similarity}
\end{table}

\subsection{Ablation Study}
\label{sec:ablation-study}

To thoroughly evaluate the effectiveness of different components of \modelname, we conduct ablation studies on three different aspects: formulations of prompt distribution, inference strategies, and choices of image encoder. Due to the page limit, our comparison against VRP-SAM with the same number of learnable parameters and more choices of image encoder, is presented in Appendix \hyperref[sec:quant-result-appendix]{Section~\ref{sec:quant-result-appendix}}
.

\par \textbf{Formulations of Prompt Distribution.} As discussed in \hyperref[sec:vpe]{Section~\ref{sec:vpe}}, a straightforward approach to modeling the prompt distribution $P(\bm{z}|I_r, M_r, I_t)$ is to assume it follows a multivariate Gaussian distribution. Consequently, we perform an ablation study under this assumption, applying the reparameterization trick inspired by VAE~\cite{vae}. As shown in \autoref{tab:ablation-params}, with different $K$ and $\nu$, \modelname\ with Gaussian prompt distributions consistently under-perform compared to t-distributions. This result highlights that the heavy-tailed nature of the t-distribution encourages a more robust mean prompt by enforcing a larger 4th-order curvature penalty, which in turn significantly enhances our testing performance on novel classes.

\begin{table}[ht!]
\renewcommand\arraystretch{0.4}
\centering
\caption{Ablation study on Gaussian prompt distribution. $K$ denotes the number of Monte-Carlo samples, and $\nu$ represents the degrees of freedom of multivariate t-distribution.}
\setlength{\tabcolsep}{0.4\tabcolsep}
\resizebox{0.9\linewidth}{!}{
\begin{tabular}{c c c c  c c c c}
\toprule
\multirow{2}{*}{Method} & \multicolumn{2}{c}{Parameters} & \multicolumn{1}{c}{PASCAL-5$^i$} & \multirow{2}{*}{Method} & \multicolumn{2}{c}{Parameters} & \multicolumn{1}{c}{PASCAL-5$^i$} \\
\cmidrule(lr){4-4}
\cmidrule(lr){8-8}
& K & $\nu$ & Means & & K & $\nu$ & Means \\
\midrule
\multirow{4}{*}{\modelname} & 10 & 5 & \best{72.04} & \multirow{4}{*}{\makecell[c]{\modelname \\ w/ Gaussian}} & 10 & - & 71.41 \\
& 10 & 3 & 71.55 & & & & \\
& 15 & 5 & 71.42 & & 15 & - & 71.01 \\
& 15 & 3 & 71.02 & & & & \\
\bottomrule
\end{tabular}%
}
\label{tab:ablation-params}
\end{table}

\textbf{Choices of Image Encoder.} Image encoder, as part of the prompt encoder, plays an important role in generating accurate prompt embeddings. A more powerful image encoder can generate image embeddings with more accurate semantics, hence leading to better prompt embeddings. To evaluate the generalizability of our method, we use a powerful self-supervised image encoder, the pre-trained DINO-v2~\cite{dinov2} ViT-B to extract visual features. As shown in \autoref{tab:ablation-dinov2}, we can see that DINOv2 indeed can significantly improve the mIOU for both VRP-SAM and our method, compared with ResNet-50. More importantly, we achieve a higher mIoU compared to VRP-SAM when both methods utilize DINOv2 as the image encoder. This result highlights the strong generalization capability of our variational prompt encoder across different image encoders.

\begin{table}[ht!]
\renewcommand\arraystretch{0.8}
\centering
\caption{Ablation study on different image encoders.}
\setlength{\tabcolsep}{\tabcolsep}
\resizebox{0.85\linewidth}{!}{%
\begin{tabular}{c  c ccccc}
\toprule
\multirow{2}{*}{Methods} & \multirow{2}{*}{\makecell[c]{Image\\Encoder}}  &\multicolumn{5}{c}{COCO-20$^i$}\\
\cmidrule(lr){3-7}
& & F-0 & F-1 & F-2 & F-3 & Mean\\
\midrule
\multirow{2}{*}{VRP-SAM} & ResNet-50 &47.02 & 54.24 & 59.91 & 51.95 & 53.28 \\
 & DINOv2 ViT-B/14 & 53.68 & 59.74 & 60.24 & 58.96 & 58.15 \\

\midrule
\ourrow 
 & ResNet-50 & 48.74 & 55.55 & 60.72 & 53.49 & 54.63\\
\ourrow
\cc{\multirow{-2}{*}{\modelname}} & DINOv2 ViT-B/14 & \best{54.49} & \best{60.57} & \best{61.81} & \best{59.8} & \best{59.16} \\
 
\bottomrule
\end{tabular}%
}
\label{tab:ablation-dinov2}
\end{table}

\textbf{Inference Strategies}. During inference, the predicted mean prompt $\bm{\hat\mu}_{\bm{z}}$ is employed to generate the predicted mask, since the robustness of $\bm{\hat\mu}_{\bm{z}}$ can be guaranteed due to the existence of margin between $\bm{\hat\mu}_{\bm{z}}$ and the boundary of target prompt region $\mathcal{R}_{I_r,M_r,I_t}$. However, it is still interesting to see the effectiveness of utilizing randomly sampled prompts from the learned prompt distribution during inference. Therefore, we experimented with 5 different inference strategies, in which we sample $K$ prompts and merge their corresponding $K$ masks in 5 different ways. Specifically, they merge either the logit masks or the binary masks given a threshold of 0.5. From \autoref{tab:ablation-infer}, all the inference strategies with different ways of merging $K$ masks achieve comparable results to using only the mask prompted by the $\bm{\hat\mu}_{\bm{z}}$. This indicates that our learned multivariate prompt distribution spans over the target prompt region.

\begin{table}[ht!]
\centering
\caption{Ablation study on different inference strategies. Their mIOU on F-0 of PASCAL-5$^i$ is presented.}
\setlength{\tabcolsep}{2.5\tabcolsep}
\resizebox{0.8\linewidth}{!}{
\begin{tabular}{c  c c }
\toprule
Methods & Inference Strategies  & mIOU \\
\midrule
\ourrow \cc{\modelname} & \cc{mean prompt only} & \best{75.26}\\ 
\midrule
\multirow{5}{*}{\modelname} & \makecell[c]{max of $K$ logit masks} & 74.69\\ 
 & \makecell[c]{mean of $K$ logit masks} & 75.16\\ 
 & \makecell[c]{max of $K$ binary masks} & 74.67\\ 
 & \makecell[c]{mean of $K$ binary masks} & 74.92\\ 
 & \makecell[c]{majority vote of $K$ binary masks} & 75.05\\ 
\bottomrule
\end{tabular}%
}
\label{tab:ablation-infer}
\vspace{-12pt}
\end{table}
\section{Conclusion}
\label{sec:conclusion}

This paper presents \modelname, a novel probabilistic prompt generation method that significantly enhances the robustness and zero-shot segmentation capability of SAM-based visual reference segmentation methods. By introducing a variational prompt encoder to learn a multivariate prompt distribution, we address the shortcomings of less robust prompts in existing approaches and thus consistently generate stable, high-quality masks. Through comprehensive experiments, \modelname\ demonstrates superior performance over SOTA methods on the Pascal-5$^i$ and COCO-20$^i$  datasets, highlighting its potential as a reliable and effective solution for visual reference segmentation. Our findings underscore the importance of probabilistic prompt generation approaches in prompt-based segmentation and pave the way for future research in this domain.

\clearpage

\setcounter{page}{1}
\maketitlesupplementary

\addcontentsline{toc}{section}{Supplementary Material} 
\renewcommand{\thepart}{} 
\renewcommand{\partname}{} 

\renewcommand{\numberline}[1]{#1\quad}
\part{Appendix} 
\parttoc

\section{Theoretical Analysis of \modelname\ and VRP-SAM}
In this section, we present a formal analysis that reveals a deep connection between variational optimization, noise injection, and regularization. As already explained in \autoref{sec:vpe}, by using the reparameterization trick, variational optimization is accomplished by adding noise to the prompt embeddings during training. In the subsequent section, we demonstrate that adding noise to the prompt embeddings is mathematically equivalent to incorporating a regularization term that penalizes the Laplacian of the loss function. This equivalence not only provides a rigorous justification for our method but also elucidates how the induced flatness in the loss landscape enhances the robustness and generalization of the model.

\subsection{Equivalence of Noise Injection and Laplacian Regularization}
\label{sec:equ-noise-lap}

\begin{proposition}
\label{prop:laplacian}
Let \( f : \mathbb{R}^n \to \mathbb{R} \) be a twice continuously differentiable function and let \(\epsilon \in \mathbb{R}^n\) be an i.i.d. distributed random noise vector satisfying
\[
\mathbb{E}[\epsilon] = 0 \quad \text{and} \quad \mathbb{E}[\epsilon\,\epsilon^T] = \sigma^2 I,
\]
where \(I\) is the \(n \times n\) identity matrix and \(\sigma > 0\) is sufficiently small. Then, for any point \( z \in \mathbb{R}^n \),
\[
\mathbb{E}_{\epsilon}\Big[f(z + \epsilon)\Big] = f(z) + \frac{\sigma^2}{2}\Delta f(z) + O(\sigma^3),
\]
where the Laplacian \(\Delta f(z)\) is defined as
\[
\Delta f(z) = \sum_{i=1}^n \frac{\partial^2 f}{\partial z_i^2}(z).
\]
\end{proposition}

The formal proof of \autoref{prop:laplacian} can be found in \autoref{proof:laplacian}.

\begin{corollary}
\label{coro:laplacian}
Assuming the \(O(\sigma^3)\) term is negligible, minimizing \(\mathbb{E}_{\epsilon}[f(z+\epsilon)]\) is equivalent to minimizing
\[
f(z) + \frac{\sigma^2}{2}\Delta f(z).
\]
\end{corollary}

\textbf{Mapping \autoref{coro:laplacian} to Variational Prompt Distribution Optimization in ProSAM.} In ProSAM, we optimize a variational prompt distribution \( q_\phi(z \mid I_r, M_r, I_t) \) using the reparameterization trick, where the sampled prompt embedding is expressed as 
\[
z = \mu_z + \epsilon,
\]
with \(\epsilon\) being a noise vector satisfying 
\[
\mathbb{E}[\epsilon]=0 \quad \text{and} \quad \mathbb{E}[\epsilon\,\epsilon^T] = \sigma^2 I.
\]
The segmentation loss is defined as 
\[
\mathcal{L}\big(f_S^M(z, F_S^I(I_t)), M_t\big),
\]
which measures the deviation between the predicted mask \( f_S^M(z, F_S^I(I_t)) \) and the ground truth mask \( M_t \). By applying \autoref{coro:laplacian} to this loss function, we obtain
\begin{align}
    &\mathbb{E}_{\epsilon}\Big[\mathcal{L}\big(f_S^M(z, F_S^I(I_t)), M_t\big)\Big] \\
\approx\ &\resizebox{0.83\linewidth}{!}{$\mathcal{L}\big(f_S^M(\mu_z, F_S^I(I_t)), M_t\big) + \frac{\sigma^2}{2}\Delta \mathcal{L}\big(f_S^M(\mu_z, F_S^I(I_t)), M_t\big)$}.
\end{align}
This shows that minimizing the expected loss over the noisy prompt embeddings is equivalent to minimizing the standard segmentation loss plus an additional regularization term that penalizes the Laplacian (i.e., the curvature) of the loss with respect to the prompt embedding \(z\) at \(z=\mu_z\). 

\subsection{The Robustness of \modelname\ by Penalizing Laplacian}
\label{sec:prosam-robust}

In this section, we explain why penalizing the Laplacian of the loss enhances the robustness of \modelname. Following the conclusion in \autoref{sec:equ-noise-lap}, injecting noise into the prompt embeddings during training is more than just a method for sampling from a variational distribution---it acts as an implicit regularizer that penalizes the Laplacian of the loss. Near local minima, small Laplacian indicates lower curvature, which is characterized by the Hessian matrix 
\[
\nabla^2 \mathcal{L}(\mu_z)
\]
of the loss function \(L\) with respect to the prompt embedding \(z\), evaluated at the mean prompt \(\mu_z\). The overall curvature is then quantified by the trace of the Hessian, namely the Laplacian,
\[
\Delta \mathcal{L}(\mu_z) =
\operatorname{Tr}\big(\nabla^2 \mathcal{L}(\mu_z)\big) = \sum_{i=1}^{n} \lambda_i,
\]
where \(\lambda_i\) are the eigenvalues of \(\nabla^2 \mathcal{L}(\mu_z)\). Near a local minimum, where the segmentation loss is minimized, the loss function is typically convex or locally convex, ensuring that all eigenvalues satisfy \(\lambda_i \geq 0\). Consequently, the Laplacian $\Delta \mathcal{L}(\mu_z)$ is nonnegative. Under this constraint, minimizing the Laplacian strictly leads to lower overall curvature. Following the conclusion from \autoref{sec:equ-noise-lap}, when noise \(\epsilon\) with variance \(\sigma^2\) is added, the Laplacian is implicitly penalized, which effectively encourages the optimization process to favor flat minima over high curvature regions. For ProSAM, this is crucial because a flat loss landscape implies that the predicted mean prompt \(\mu_z\) is robust to small perturbations, thereby enhancing the stability and generalization of segmentation performance, particularly on novel objects.

\subsection{The Limitation of VRP-SAM Without Laplacian Regularization}
\label{sec:vrp-drawback}

In VRP-SAM, the prompt encoder is optimized solely by minimizing the segmentation loss:
\[
\mathcal{L}(\mu_z) = \mathcal{L}\big(f_S^M(\mu_z, F_S^I(I_t)), M_t\big),
\]
where \(\mu_z\) is the learned prompt embedding, \(f_S^M\) denotes the SAM mask decoder, \(F_S^I(I_t)\) is the image feature extraction, and \(M_t\) represents the ground truth mask. This objective ensures that the generated mask is close to the target mask but does not explicitly encourage the embedding \(\mu_z\) to reside in the low-curvature area of the target prompt region \(R_{I_r, M_r, I_t}\). As a result, the learned embedding may end up in an area where the loss function exhibits high curvature.

As explained in \autoref{sec:prosam-robust}, the curvature at the embedding \(\mu_z\) is characterized by the Hessian \(\nabla^2 \mathcal{L}(\mu_z)\) of the loss function, and its trace, the Laplacian $\Delta \mathcal{L}(\mu_z)$ can be large if \(\mu_z\) is near a boundary or a sharp region of the loss landscape. Without a regularization term that penalizes this curvature—such as the additional term \(\frac{1}{2}\sigma^2\, \Delta \mathcal{L}(\mu_z)\) obtained via noise injection—the optimizer is not explicitly guided to find flatter regions. Consequently, small perturbations in the embedding can lead to significant increases in loss, making the model more sensitive to noise and less robust. This sensitivity is particularly problematic when segmenting novel objects, where the embedding must generalize well to unseen variations. Hence, the absence of Laplacian regularization in VRP-SAM can result in unstable prompt embeddings and degraded segmentation performance.

\subsection{Mathematical Proofs}
\label{proof:laplacian}

\begin{proof}
\textbf{Step 1. Taylor Expansion} \\
Since \( f \) is twice continuously differentiable, we can write the second-order Taylor expansion of \( f(z+\epsilon) \) about the point \( z \):
\begin{equation}
\label{equ:taylor-exp}
f(z+\epsilon) = f(z) + \nabla f(z)^T \epsilon + \frac{1}{2}\epsilon^T H_f(z) \epsilon + R(\epsilon)
\end{equation},
where 
\begin{itemize}
    \item \(\nabla f(z)\) is the gradient of \(f\) at \(z\),
    \item \(H_f(z)\) is the Hessian matrix of \(f\) at \(z\),
    \item \(R(\epsilon)\) is a remainder term of order \(O(\|\epsilon\|^3)\).
\end{itemize}

\textbf{Step 2. Taking the Expectation} \\
Taking the expectation with respect to \(\epsilon\), we obtain:
\[
\mathbb{E}_{\epsilon}\Big[f(z+\epsilon)\Big] = \mathbb{E}_{\epsilon}\left[ f(z) + \nabla f(z)^T \epsilon + \frac{1}{2}\epsilon^T H_f(z) \epsilon + R(\epsilon) \right].
\]
Since \(f(z)\) is constant and \(\mathbb{E}_{\epsilon}[\epsilon] = 0\), we have:
\begin{equation}
\label{equ:taylor}
\mathbb{E}_{\epsilon}\Big[f(z+\epsilon)\Big] = f(z) + \frac{1}{2}\mathbb{E}_{\epsilon}\Big[\epsilon^T H_f(z) \epsilon\Big] + O(\sigma^3).
\end{equation}

\textbf{Step 3. Evaluating the Quadratic Term} \\
Express the quadratic form as:
\[
\epsilon^T H_f(z) \epsilon = \sum_{i=1}^n\sum_{j=1}^n H_f(z)_{ij} \,\epsilon_i\,\epsilon_j.
\]
Taking the expectation, we have:
\[
\mathbb{E}_{\epsilon}\Big[\epsilon^T H_f(z) \epsilon\Big] = \sum_{i=1}^n\sum_{j=1}^n H_f(z)_{ij} \,\mathbb{E}_{\epsilon}[\epsilon_i\,\epsilon_j].
\]
Given that \(\mathbb{E}_{\epsilon}[\epsilon_i\,\epsilon_j] = \sigma^2\) if \(i = j\) and \(0\) otherwise, it follows that:
\[
\mathbb{E}_{\epsilon}\Big[\epsilon^T H_f(z) \epsilon\Big] = \sigma^2 \sum_{i=1}^n H_f(z)_{ii} = \sigma^2\, \operatorname{Tr}(H_f(z)).
\]
Recall that the Laplacian of \(f\) is defined as:
\[
\Delta f(z) = \operatorname{Tr}(H_f(z)) = \sum_{i=1}^n \frac{\partial^2 f}{\partial z_i^2}(z).
\]

\textbf{Step 4. Final Expression} \\
Substituting the evaluated quadratic term into our expectation, we obtain:
\[
\mathbb{E}_{\epsilon}\Big[f(z+\epsilon)\Big] = f(z) + \frac{1}{2}\sigma^2\,\Delta f(z) + O(\sigma^3).
\]
This completes the proof.
\end{proof}

\subsection{Advantage of Student-$t$ over Gaussian}
\label{sec:t-advantage}

Following \autoref{equ:taylor}, we observed that the second‐order Taylor terms of $\mathbb{E}[L(z+\epsilon)]$ are identical for any zero‐mean noise with covariance $\sigma^2I$. Third‐order terms also vanish because the noise distributions are symmetric and have zero odd moments. The distinction appears first in the fourth‐order term, which depends on the fourth central moment

$$
  m_4 = \mathbb{E}[\epsilon_i^4].
$$

For a Gaussian $\mathcal{N}(0,\sigma^2)$, one has

$$
  m_4^{\mathcal N} = 3\,\sigma^4,
$$

whereas for a Student–$t$ with $\nu>4$ degrees of freedom, scaled to variance $\sigma^2$,

$$
  m_4^{t} = \frac{3\,\nu}{\nu-4}\,\sigma^4 > 3\,\sigma^4.
$$

Recalling that the fourth‐order correction in the expected loss is

$$
  \frac{1}{24}\sum_{i,j,k,\ell}
  \frac{\partial^4L}{\partial z_i\partial z_j\partial z_k\partial z_\ell}(z)
  \,\mathbb{E}[\epsilon_i\epsilon_j\epsilon_k\epsilon_\ell],
$$

and that only index‐pairings $(i=j=k=\ell)$ and $(i=j\neq k=\ell)$ survive, the larger $m_4$ of the Student–$t$ directly amplifies the contribution

$$
  \frac{m_4}{24}\sum_i Q_{iiii}
  +\;
  \frac{3\,\sigma^4}{24}\sum_{i\neq j}Q_{iijj},
$$

where $Q_{ijkl}=\frac{\partial^4L}{\partial z_i\partial z_j\partial z_k\partial z_\ell}$. Consequently, Student–$t$ noise imposes a strictly stronger fourth‐order “push” against high curvature than Gaussian noise, driving the mean prompt deeper into flatter regions of the loss landscape and yielding greater empirical robustness.

\begin{figure*}[!ht]
  \centering
  \includegraphics[width=\linewidth]{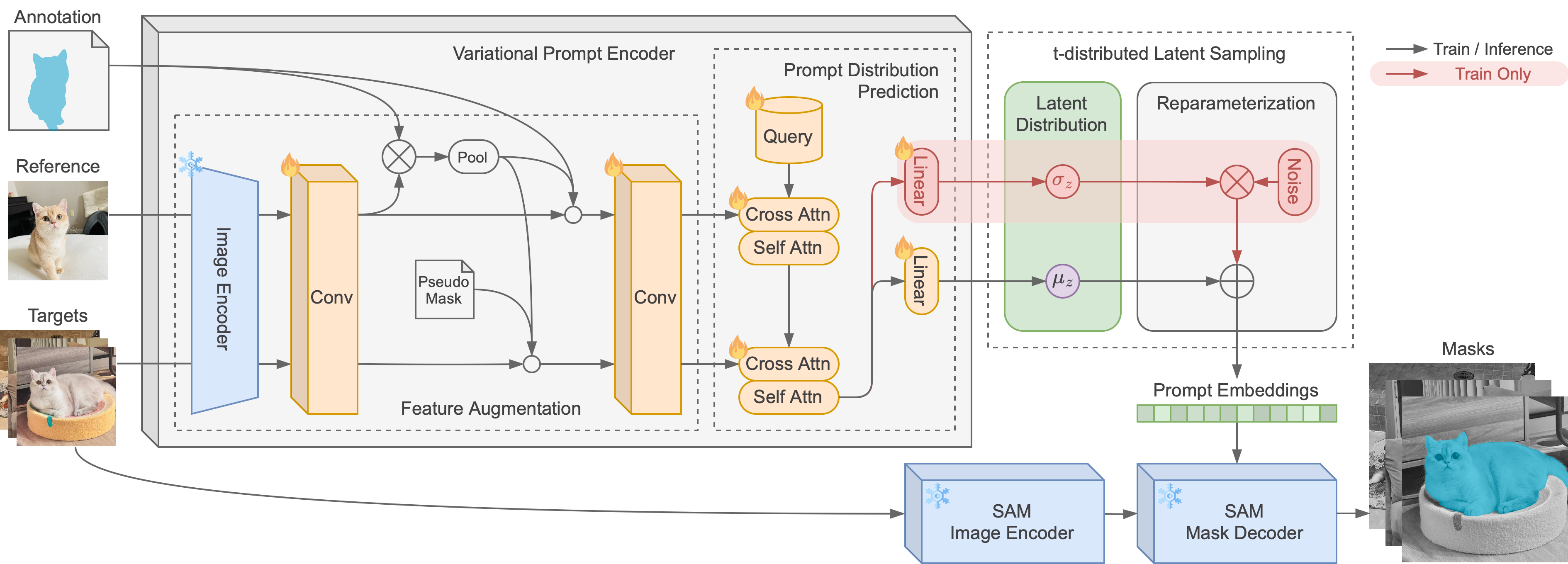}
  \caption{The detailed model architecture of \modelname. The only trainable module in \modelname\ is the variational prompt encoder, which is composed of two components: feature augmentation and prompt distribution prediction. Specifically, the feature augmentation aims to extract the enhanced reference and target feature to guide the learning of prompt distribution. The prompt distribution prediction module is responsible for predicting the variational prompt distribution to guide the SAM in mask generation for the target images.}
  \label{fig:architecture}
\end{figure*}

\section{Model Architecture}
\label{sec:model-archi}
In this section, the model architecture of our variational prompt encoder is described in detail. For a fair and straightforward comparison, our variational prompt encoder closely follows the model architecture of VRP-SAM~\cite{vrpsam} (see \autoref{sec:vrpsam}). As shown in \autoref{fig:architecture}, our variational prompt encoder is composed of two major components: feature augmentation and prompt distribution prediction. 

\subsection{Feature Augmentation} 

In this component, the visual features of the reference image $I_r$ and the target image  $I_t$ are augmented and enhanced with reference annotations $M_r$. First, both the reference image $I_r$ and the target image $I_t$ are encoded into $F_{I_r}$ and $F_{I_t}$ using a frozen pre-trained image encoder $f_I$ followed by a learnable pointwise convolutional layer. To obtain a reference annotation embedding $F_{M_r}$, the reference image embedding $F_{I_r}$ within the annotated region $M_r$ is fed into an average pooling layer. Next, $F_{M_r}$ is concatenated with both the reference image embedding $F_{I_r}$ and the reference annotation $M_r$ in a pointwise manner, and then transformed by another pointwise convolutional layer to produce the final enhanced reference feature $F_r^v$. To obtain the enhanced target feature $F_t^v$, a pseudo-mask of target image $M_t^{pseudo}$ is generated by evaluating the pixel-wise similarity map through the comparison of high-level features of reference and target image. Then, the similarity map is normalized into [0,1] and serves as the pseudo mask $M_t^{pseudo}$ for the target image. Similarly, the enhanced target feature $F_t^v$ is obtained by transforming a concatenation of $M_t^{pseudo}$, $F_{M_r}$ and $F_{I_t}$ with a learnable pointwise convolution layer.

\subsection{Prompt Distribution Prediction} 

Given the enhanced reference feature $F_r^v$ and target feature $F_t^v$, a variational prompt distribution $q_{\phi}(\bm{z}|I_r,M_r,I_t)$ is predicted via attention mechanisms. First, a set of learnable queries $Q \in \mathbb{R}^{m \times c}$ is initialized and interacted with the reference feature $F_r^v$ through a cross-attention layer and a self-attention layer, to generate query vectors $Q_r' \in \mathbb{R}^{m \times c}$ containing information about the object to be segmented. These query vectors $Q_r'$ then interact with the target feature $F_t^v$ via another cross-attention layer and a subsequent self-attention layer to produce the prompt features $Q_t' \in \mathbb{R}^{m \times c}$. Finally, two linear transformation heads are employed to predict the mean $\bm{\hat\mu}_{z}$ and standard deviation $\bm{\hat\sigma}_{\bm{z}}$ of the variational prompt distribution $q_{\phi}(\bm{z}|I_r,M_r,I_t)$, respectively. For the quantitative comparison against VRP-SAM with two linear layers appended at the end of prompt encoder, please refer to \autoref{sec:vrpsam-2linear}.

\section{Additional Verification Study}
\label{sec:more-verify-study}

In addition to the studies presented in \autoref{sec:verify-study}, we also designed a verification study that does not require training a deep learning model, allowing for a direct comparison between the underlying principles of the variational and non-variational prompt encoders. 

Specifically, given the pre-trained SAM mask decoder and SAM image encoder, we learn the prompt embedding or variational prompt distribution via gradient descent for a given object (i.e., image-mask pair).  To learn the prompt embedding via gradient descent, the gradient $\frac{\partial \mathcal{L}}{\partial \bm{z}}$ will be computed and used to directly update $\bm{z}$, which is treated as a parameterized vector rather than being predicted by the prompt encoder. Similarly, to learn the multivariate prompt distribution, the gradient $\frac{\partial \mathcal{L}}{\partial \bm{\hat\mu}_{\bm{z}}}$ and  $\frac{\partial \mathcal{L}}{\partial  \bm{\hat\sigma}_{\bm{z}}}$ will be utilized to update the parameterized vector $\bm{\hat\mu}_{\bm{z}}$ and $\bm{\hat\sigma}_{\bm{z}}$. Essentially, the learned prompt embedding reflects the fundamental principles of a non-variational prompt encoder (e.g., VRP-SAM), while the learned prompt distribution captures the core principles of a variational prompt encoder (e.g., \modelname).

For the experiment of learning prompt embedding, we ran 200 experiments to generate 200 prompt embeddings. The initial prompt embeddings are randomly drawn from the normal distribution with a mean of 0 and a standard deviation of 25, and the loss function is formulated as the same loss function as VRP-SAM (see \autoref{equ:vrpsam-loss}). For the experiment of learning variational prompt distribution, we learn a single multivariate prompt distribution following the same formulation and reparameterization trick in \autoref{equ:prompt-dist} and \autoref{equ:reparam-trick} via gradient descent, given the loss function presented in \autoref{equ:objective}. For both experiments, the gradient descent optimization process is stopped when the loss value does not improve by more than 0.0003 over 100 consecutive epochs. For other experimental settings not mentioned above, we follow the same practice as in \autoref{sec:experiment-setup}.

\begin{figure*}[ht!]
  \centering
  \includegraphics[width=0.95\linewidth]{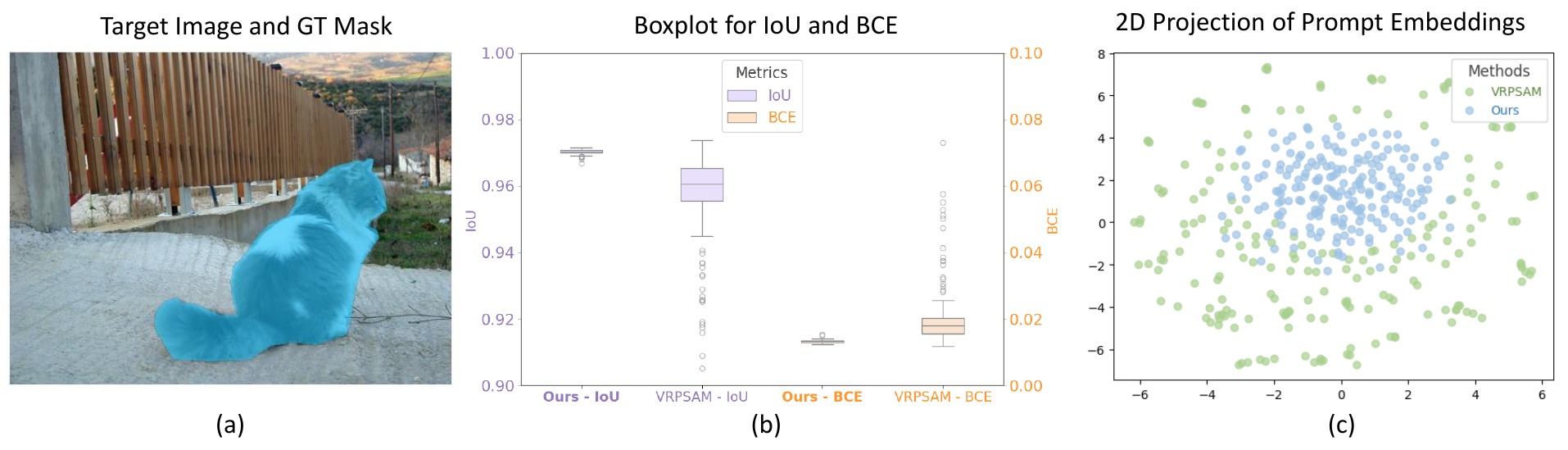}
  \caption{The visualization of the learned prompt embeddings by VRP-SAM and our method through gradient descent. For a sample image from COCO-20$^i$ presented in Figure (a), we analyze the generated prompt embeddings and their associated mask predictions in Figure (b) and (c). Specifically, in Figure (b), the IoU (left y-axis) and BCE (right y-axis) are computed between the predicted masks and the ground-truth mask. In Figure (c), the 2D projection of prompt embeddings via t-SNE is visualized.}
  \label{fig:embedding-plot}
  \vspace{-5pt}
\end{figure*}

The visualization results on a sample image are presented in \autoref{fig:embedding-plot}. First, from \autoref{fig:embedding-plot}(b), we can see that both VRP-SAM and \modelname\ are able to generate faithful prompts with IoU close to 0.96 and BCE close to zero. Notably, the prompt embeddings sampled from our variational prompt distribution consistently perform better than at least 75\% of 200 prompt embeddings learned by VRP-SAM, with higher IoU and lower BCE value. From the scatter plot of projected prompt embeddings via t-SNE~\cite{tsne} in \autoref{fig:embedding-plot}(c), we can observe that the prompts sampled from our variational prompt distribution are clearly clustered in the center, while solely learning a single observation of prompt embeddings lie in the boundary of our variational prompt distribution. This observation assures that the proposed variational prompt encoder can indeed produce more robust prompts that are closer to the center of the target prompt region  $\mathcal{R}_{I_r,M_r,I_t}$, compared with the non-variational prompt encoder employed by VRP-SAM.

\section{Additional Quantitative Evaluations}
\label{sec:quant-result-appendix}

To thoroughly assess the effectiveness of \modelname, more quantitative evaluations have been conducted and presented here due to the page limit. First, we analyze our confusion matrix compared with VRP-SAM confusion matrix in \autoref{sec:conf-matrix} for a detailed comparison. Secondly, to conduct a fair comparison against VRP-SAM with the same number of parameters as ours, we present an ablation study on VRP-SAM with two linear layers appended in \autoref{sec:vrpsam-2linear}. Additionally, an ablation study on different inference strategies has been conducted in \autoref{sec:ablation-study}. Lastly, we present more quantitative results on different choice of image encoder in \autoref{sec:image-encoder}. Again, similar to the results presented in \hyperref[sec:result]{Section~\ref{sec:result}}, the experimental results presented here for both VRP-SAM and our method are conducted under identical experimental settings to ensure a fair comparison.

\subsection{Confusion Matrix Comparison with VRP-SAM} 
\label{sec:conf-matrix}
As demonstrated in \autoref{sec:quant-result}, we have outperformed the state-of-the-art method VRP-SAM on both COCO-20$^i$ and PASCAL-5$^i$ (see \autoref{tab:fm-result}), and surpassed VRP-SAM under the significant domain shift from COCO-20$^i$ to PASCAL-5$^i$ (see \autoref{tab:domain_shift}). The question is whether \modelname\ will potentially suffer from a greater false negative rate (FNR) while predicting a mean prompt that is more aligned with the center of target prompt region $\mathcal{R}_{I_r,M_r,I_t}$ (defined in \autoref{sec:vpe}). The detailed evaluations of \modelname\ and VRP-SAM on True Positive Rate (TPR), True Negative Rate (TNR), False Positive Rate (FPR), and False Negative Rate (FNR) have been presented in \autoref{tab:conf-matrix}. As you can see, for every fold in PASCAL-5$^i$, \modelname\ can obtain a higher TPR and TNR in terms of pixel-level accuracy, while reducing the FPR and FNR systematically.

\begin{table}[htbp!]
    \centering
    \caption{A detailed comparison of \modelname\ predictions with VRP-SAM predictions. At here, the True Positive Rate (TPR), True Negative Rate (TNR), False Positive Rate (FPR), and False Negative Rate (FNR) have been computed for both \modelname\ and VRP-SAM predictions on PASCAL-5$^i$. }
    \label{tab:conf-matrix}
    \resizebox{\linewidth}{!}{
    \begin{tabular}{c c c c c c }

\toprule

 \addlinespace[-2.5pt] \multirow{2}{*}{Methods} & \multirow{2}{*}{Metrics} & \multicolumn{4}{c}{PASCAL-5$^i$}\\
\addlinespace[-2pt] 
\cmidrule(lr){3-6}

\addlinespace[-2pt] 
 & & F-0 & F-1 & F-2 & F-3 \\
 \addlinespace[-2pt] 
\midrule
\addlinespace[1pt] 
\modelname  & \multirow{2}{*}{TPR(\%)} & \best{78.47} & \best{68.97} & \best{66.97} & \best{66.37}   \\
VRPSAM & & 78.14 & 68.21 & 65.88 & 64.39   \\
\addlinespace[-2pt] 
\midrule
\addlinespace[1pt] 
\modelname  & \multirow{2}{*}{TNR(\%)} & \best{11.55} & \best{19.91} & \best{17.82} & \best{16.74}   \\
VRPSAM  & & 11.54 & 19.89 & 17.22 & 16.67   \\
\addlinespace[-2pt] 

\midrule
\addlinespace[1pt] 
\modelname  & \multirow{2}{*}{FPR(\%)} & \best{9.21} & \best{9.5} & \best{12.75} & \best{14.13} \\
VRPSAM  & & 9.53 & 10.25 & 13.84 & 16.12 \\
\addlinespace[-2pt] 
\midrule
\addlinespace[1pt]
\modelname  & \multirow{2}{*}{FNR(\%)} & \best{0.78} & \best{1.62} & \best{2.46} & \best{2.75} \\
VRPSAM  & & \best{0.78} & 1.64 & 3.05 & 2.82 \\
\addlinespace[-2pt]

\bottomrule
\end{tabular}
}
    
\end{table}

\subsection{VRP-SAM with Same Number of Parameters as \modelname} 
\label{sec:vrpsam-2linear}
As described in \autoref{sec:experiment-setup} and \hyperref[sec:model-archi]{Section~\ref{sec:model-archi}}, the major difference in our model architecture compared with VRP-SAM is that we append two linear layers at the end of the variational prompt encoder to predict the mean and variance of the prompt distributions. Therefore, \modelname\ has more learnable parameters resulting from these two linear layers. To conduct a fair comparison under the same number of learnable parameters, we trained a VRP-SAM with two linear layers appended at the end of their prompt encoder while keeping other experimental settings the same. From \autoref{tab:vrpsam-2linear}, we can see that appending two linear layers at the end of VRP-SAM prompt encoder fails to boost the VRPSAM performance. In other words, with the same number of learnable parameters, \modelname\ still surpasses VRP-SAM by a large margin.

\begin{table}[!htbp]
    \centering
    \caption{A quantitative comparison against VRP-SAM with the exactly same number of learnable parameters as ours. To be specific, 2 linear layers have been appended at the end of VRP-SAM prompt encoder to ensure identical model architecture as ours (see the second row below).}
    \label{tab:vrpsam-2linear}
    \resizebox{\linewidth}{!}{
    \begin{tabular}{c c c c c c }
\toprule
\addlinespace[0.5pt]
\multirow{2}{*}{Methods} & \multirow{2}{*}{Metrics} & \multicolumn{4}{c}{PASCAL-5$^i$}\\
\addlinespace[-2pt]
\cmidrule(lr){3-6}
\addlinespace[-2pt]
 & & F-0 & F-1 & F-2 & F-3 \\
 \addlinespace[-2pt]
\midrule
\addlinespace[0.5pt]
\ourrow ProSAM & \multirow{3}{*}{mIOU(\%)} & \best{75.26} & \best{77.57} & \best{70.29} & \best{65.22} \\
VRPSAM+2Linear &  & 74.04 & 76.55 & 69.71 & 63.98 \\
VRPSAM &  & 74.01 & 76.77 & 69.46 & 64.34 \\

\addlinespace[-2pt]
\bottomrule
\end{tabular}
}
\end{table}

\subsection{Choices of Image Encoder} 
\label{sec:image-encoder}

In addition to ResNet-50~\cite{resnet} and DINOv2~\cite{dinov2}, we also experimented on adopting VGG-16~\cite{vgg} as the image encoder. From \autoref{tab:ablation-vgg}, we can see that \modelname\ with VGG-16 surpasses VRP-SAM with VGG-16 for all different folds. Also, for both VRP-SAM and \modelname, the performance with VGG-16 generally performs worse than the performance with ResNet-50. This indicates that ResNet-50 can extract more accurate semantic-aware visual features and thereby enable \modelname\ to learn better prompts.

\begin{table}[ht!]
\centering
\caption{The quantitative evaluations of \modelname\ with different image encoders such as ResNet-50 and VGG-16.}
\setlength{\tabcolsep}{0.9\tabcolsep}
\resizebox{\linewidth}{!}{%
\begin{tabular}{c  c ccccc}
\toprule
\multirow{2}{*}{Methods} & \multirow{2}{*}{\makecell[c]{Image\\Encoder}}  &\multicolumn{5}{c}{PASCAL-5$^i$}\\
\cmidrule(lr){3-7}
& & F-0 & F-1 & F-2 & F-3 & Mean\\
\midrule
\multirow{2}{*}{VRP-SAM} & ResNet-50 &74.01 & 76.77 & 69.46 & 64.34 & 71.14 \\
 & VGG-16 & 69.72 & 74.74 & 67.12 & 61.84 & 68.35\\

\midrule
\ourrow 
 & ResNet-50 & \best{75.26} & \best{77.57} & \best{70.09} & \best{65.22} & \best{72.04}\\
\ourrow
\cc{\multirow{-2}{*}{\modelname}} & VGG-16 & 70.53 & 75.30 & 68.25 & 62.99 & 69.27 \\
 
\bottomrule
\end{tabular}%
}
\label{tab:ablation-vgg}
\vspace{-1pt}
\end{table}

\section{Qualitative Evaluations}
\label{sec:quail-result}
To qualitatively evaluate the effectiveness of \modelname, we first present a qualitative comparison with VRP-SAM on COCO-20$^i$ in \autoref{fig:qual-compare}, then showcase the generalizability of \modelname\ on diverse image styles in \autoref{fig:qual-general} and lastly demonstrate our capability of handling challenging cases in \autoref{fig:complex-image-results}.

\subsection{Qualitative Comparison with VRP-SAM} 

After a thorough qualitative analysis of masks generated by \modelname\ and VRP-SAM across multiple datasets, we observed a general trend: our generated masks are less prone to artifacts, such as small holes or disconnected regions, which often appear in the masks produced by VRP-SAM. 

\begin{figure*}[htbp!]
    \centering
    \includegraphics[width=\linewidth]{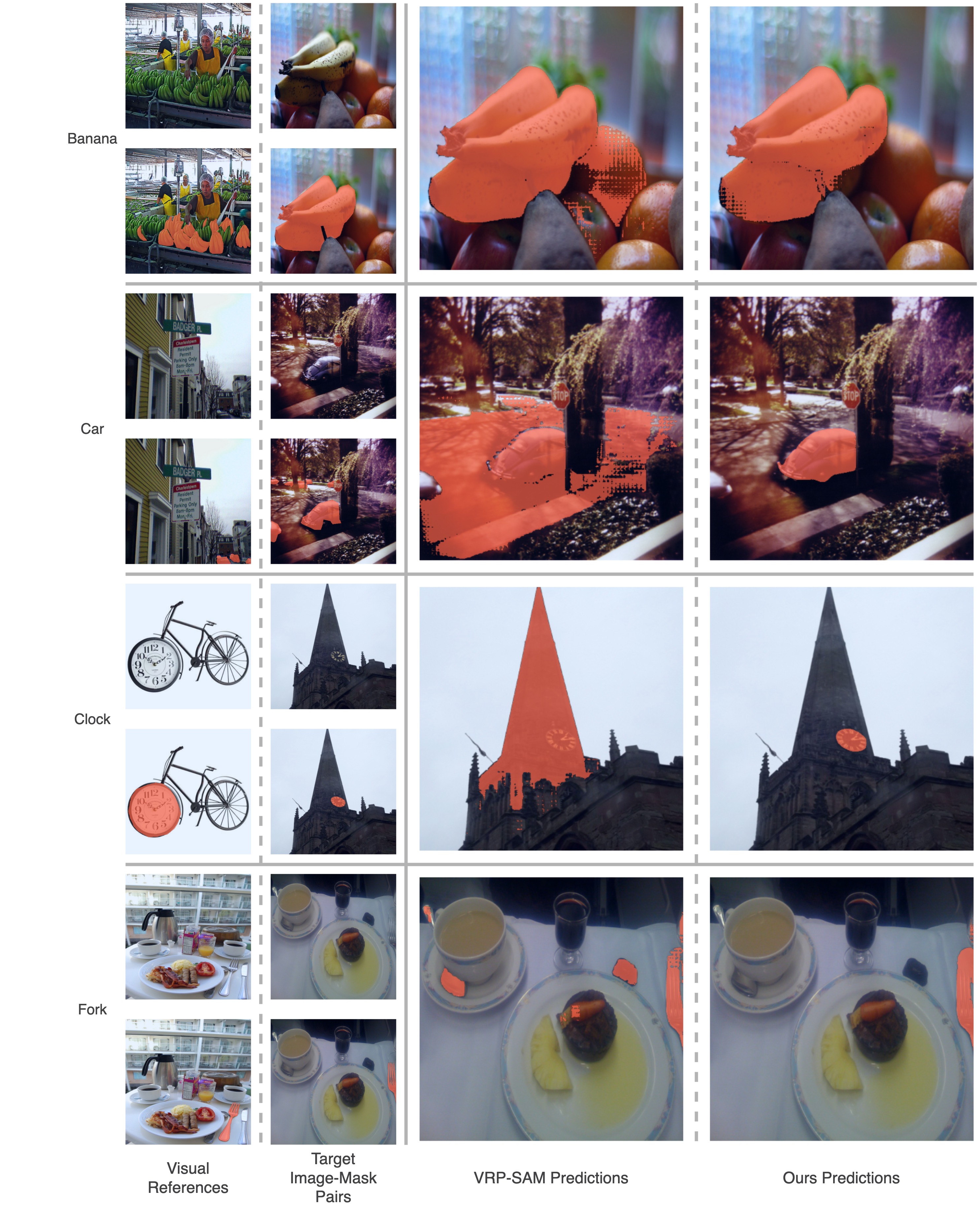}
    \caption{Qualitative comparison between VRP-SAM and \modelname\ on COCO-20$^i$.}
    \label{fig:qual-compare}
\end{figure*}

For example, in \autoref{fig:qual-compare}, VRP-SAM predictions on "car" and "banana" exhibit many small holes and pixelated artifacts in the masked region, whereas our predictions are consistently more robust with fewer pixelated artifacts. One key reason is that the mean prompts of our learned prompt distribution are more robust and precise than prompts predicted by VRP-SAM because our mean prompts are encouraged to be more closely aligned with the center of the target prompt region during the training. Thus, the masks generated by our mean prompts have higher quality. It is also interesting to see that VRP-SAM masks tend to have more false positives, which is consistent with our findings in \autoref{sec:conf-matrix}. Taking "car" and "clock" in \autoref{fig:qual-compare} as examples: VRP-SAM wrongly perceives the road as "car"; the entire spire is incorrectly predicted as "clock" by VRP-SAM. However, by taking advantage of the robustness of our predicted mean prompts, our mask predictions on "car" and "clock" are accurate and precise with much fewer false positives. For "fork", VRP-SAM not only predicts more false positives but also wrongly treats other silverware (e.g., spoon and knife) as a "fork", while we generate a more accurate mask for "fork" by leveraging a more optimal prompt encoder.

\subsection{Generalizability on Diverse Image Styles} 

\begin{figure*}[htbp!]
    \centering
    \includegraphics[width=\linewidth]{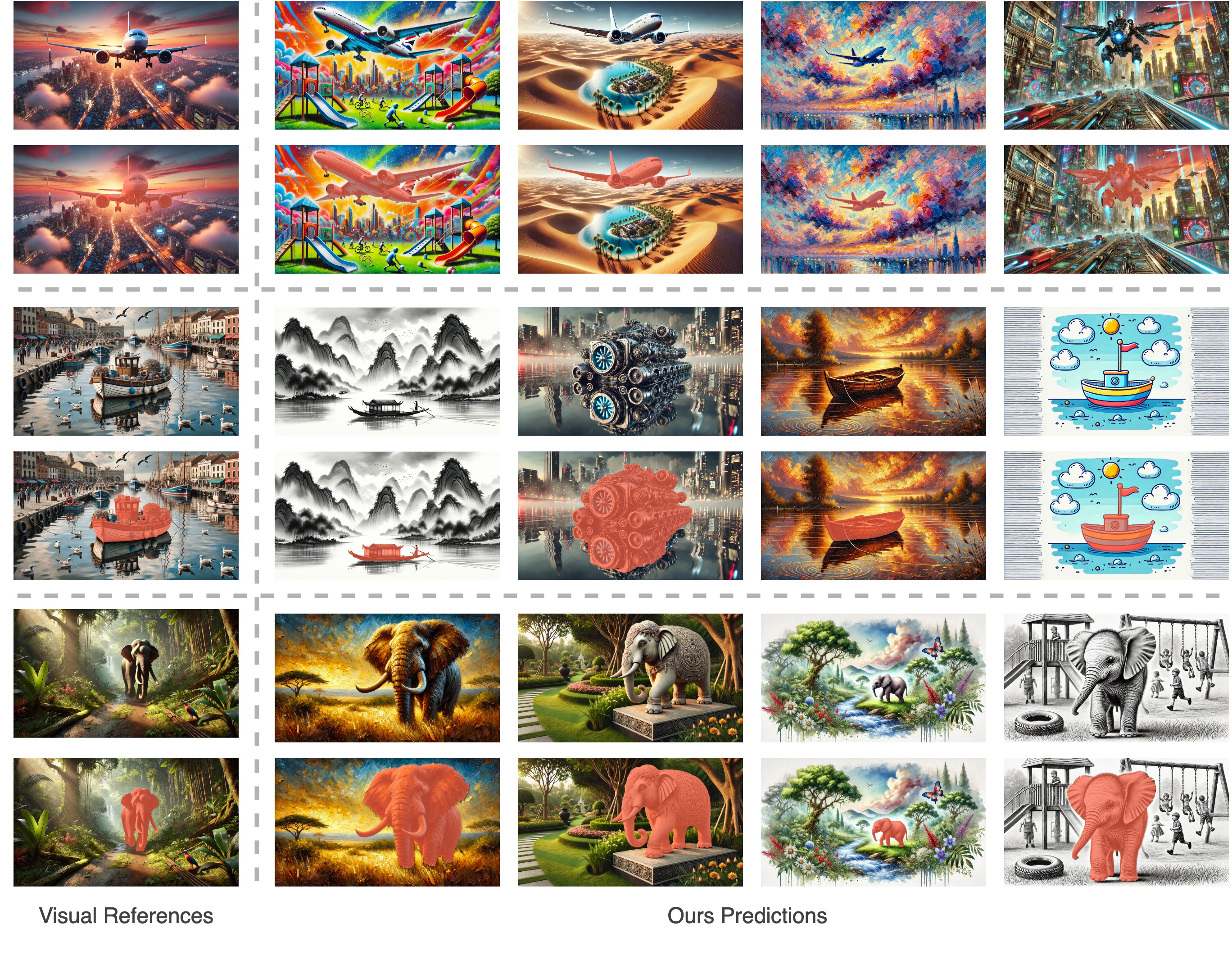}
    \caption{Qualitative results of \modelname\ (trained on COCO-20$^i$) across diverse image styles. Both the reference images and target images were collected from the internet.}
    \label{fig:qual-general}
\end{figure*}

To evaluate the generalizability of \modelname\ on images with novel and unseen styles, we conducted experiments on images featuring complex scenes and diverse styles. Specifically, both reference images and target images were collected from the internet, and the reference annotations were curated by prompting SAM with bounding boxes. As demonstrated in \autoref{fig:qual-general}, even though the model is trained on general-style images only (COCO-20$^i$), \modelname\ can consistently generate high-quality masks with precise and clean boundaries, regardless of whether the target images are artistic paintings or photorealistic scenes. The ability to maintain such performance, even across vastly different image styles, is particularly impressive, as it requires no re-training or fine-tuning of the model. This strongly highlights the zero-shot segmentation capability of \modelname\ in open-world scenarios.

\subsection{Capability of Handling Challenging Cases} 
In image segmentation, certain challenging scenarios often cause segmentation methods to fall short. One such challenge arises when target objects have irregular shapes and non-uniform boundaries, which can lead to artifacts along object edges. Another common difficulty occurs when an image contains multiple target objects, as some objects may be overlooked, either receiving no masks or being assigned low-quality masks.

\begin{figure*}
    \centering
    \includegraphics[width=\linewidth]{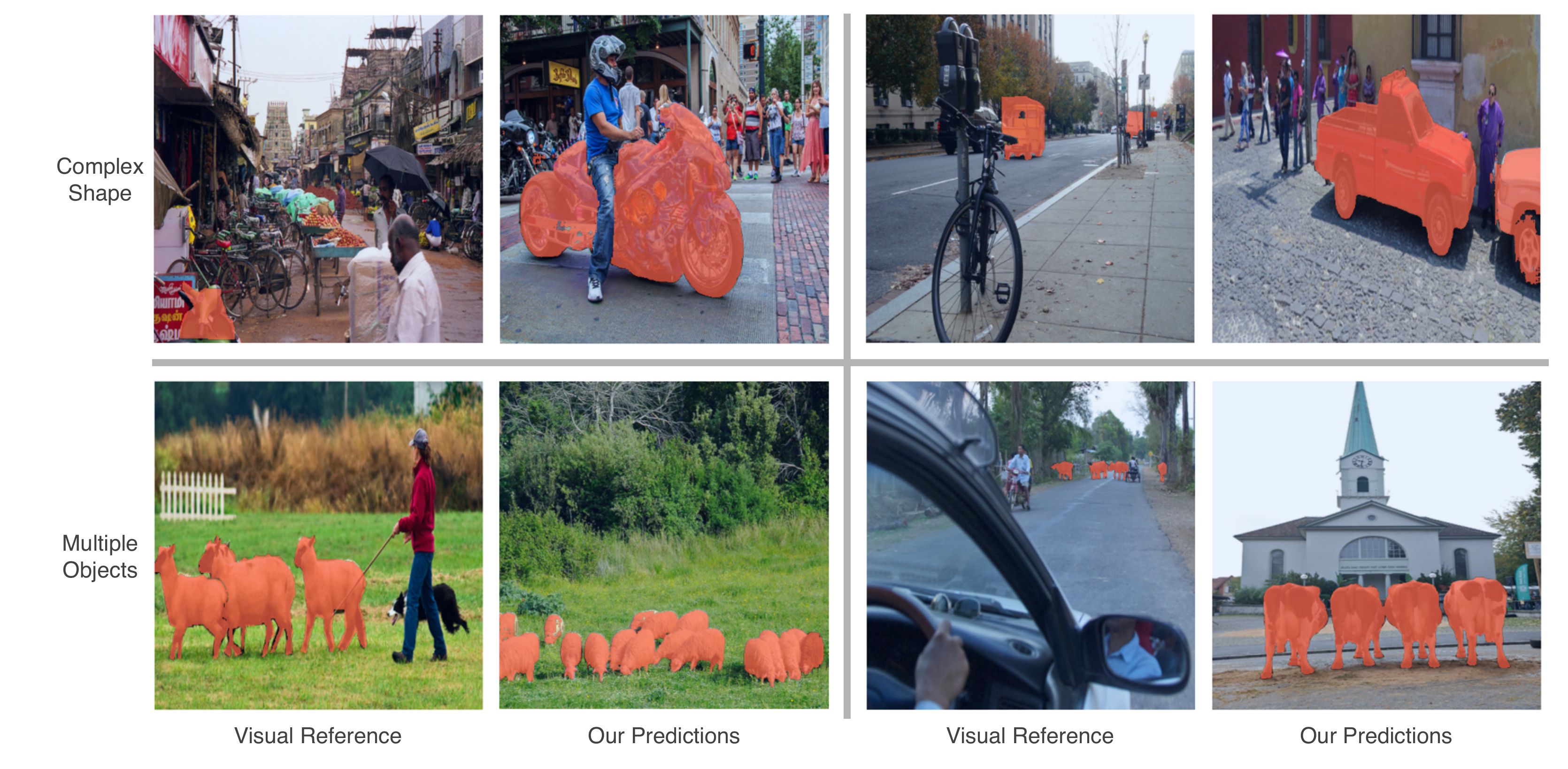}
    \caption{Qualitative results of \modelname\ on two famous challenging cases including segmenting objects with irregular shape and segmenting multiple target objects.}
    \label{fig:complex-image-results}
\end{figure*}

To better showcase our capability in understanding visual references and handling these challenges, we present qualitative results for these two scenarios in \autoref{fig:complex-image-results}. The visualization results demonstrate that \modelname\ effectively generates high-quality masks even in the presence of complex shapes and multiple target objects. A key reason behind this strong performance is that our variational prompt encoder jointly learns multiple prompt distributions to guide SAM, enabling it to capture both non-uniform object boundaries and multiple objects within a scene. For example, in \autoref{fig:complex-image-results}, even though the motorcycle has an irregular boundary, our predictions accurately capture its complexity, producing a high-quality mask.  Additionally, for the target images containing 15 goats, \modelname\ successfully detects and segments all of them, demonstrating its robustness in handling multiple target objects.

\clearpage
\clearpage

{
    \small
    \bibliographystyle{ieeenat_fullname}
    \bibliography{main}
}

\end{document}